\documentclass{article}

\usepackage[final,main]{neurips_2026}

\usepackage[utf8]{inputenc}
\usepackage[T1]{fontenc}
\usepackage{hyperref}
\usepackage{url}
\usepackage{booktabs}
\usepackage{amsfonts}
\usepackage{nicefrac}
\usepackage{microtype}
\usepackage{xcolor}
\usepackage{amsmath}
\usepackage{amssymb}
\usepackage{amsthm}
\usepackage{graphicx}
\usepackage{epstopdf}
\usepackage{multirow}
\usepackage{makecell}
\usepackage{latexsym}
\usepackage{subcaption}
\usepackage{bm}
\usepackage{colortbl}
\usepackage{tabularx}
\definecolor{ragtableblue}{HTML}{287FE0}
\definecolor{ragtabletext}{HTML}{205B98}
\usepackage[ruled,vlined,linesnumbered]{algorithm2e}
\usepackage{inconsolata}
\usepackage{wrapfig}

\newtheorem{lemma}{Lemma}

\usepackage{amsmath,amsfonts,bm}

\def\eqref#1{equation~\ref{#1}}
\def\1{\bm{1}}

\DeclareMathAlphabet{\mathsfit}{\encodingdefault}{\sfdefault}{m}{sl}
\SetMathAlphabet{\mathsfit}{bold}{\encodingdefault}{\sfdefault}{bx}{n}

\title{In-Context Optimization for Retrieval-Augmented Generation: A Gradient-Descent Perspective}

\author{
Mingchen Li\textsuperscript{\normalfont 1,6}\thanks{%
indicates equal contribution
\hspace{0.8em}\textbar\hspace{0.8em}
The code is available at
\href{https://github.com/ToneLi/ICL_RAG}{GitHub}.%
},
Jiatan Huang\textsuperscript{\normalfont 2}\footnotemark[1],
Chuxu Zhang\textsuperscript{\normalfont 2},
Liang Zhao\textsuperscript{\normalfont 3,4},
\textbf{Hong Yu}\textsuperscript{\normalfont 1,5,6}\\[6pt]
\textsuperscript{1}University of Massachusetts Amherst \qquad
\textsuperscript{2}University of Connecticut \\[3pt]
\textsuperscript{3}Emory University \qquad
\textsuperscript{4}Causal Dynamics Lab \\[3pt]
\textsuperscript{5}University of Massachusetts Lowell \qquad
\textsuperscript{6}VA Bedford Health Care
}
        
\begin{document}

\maketitle

\begin{abstract}
In-context learning has recently been linked to implicit gradient descent in linear self-attention models, suggesting that context can induce a forward-pass update. Retrieval-augmented generation (RAG) also relies on context, but retrieved documents are usually treated as static evidence rather than signals for adaptation. We study RAG as an in-context optimization process. First, we show that one linear self-attention layer can implement one gradient-descent step on a unified linearized RAG objective covering both projection-based and dot-product retrieval interfaces. This gives an exact regime where retrieval-augmented prediction and in-context optimization coincide. We use this result not as a literal model of LLM computation, but as a guide for adapting the interaction between queries and retrieved evidence. We then test the boundary of this correspondence: it remains stable under controlled linear extensions, but becomes feature-distribution dependent under nonlinear architectures. Finally, we turn this view into a lightweight method for frozen RAG LLMs. The method keeps the retriever and backbone fixed, and predicts a context-conditioned update to a generator-side evidence-use interface. Across seven QA benchmarks, two retrievers, and two frozen LLM backbones, this forward-only update improves a shared-interface baseline, transfers to held-out tasks, and approaches test-time gradient adaptation at much lower per-query cost.
\end{abstract}

\section{Introduction}

Large language models (LLMs) have achieved strong performance across many natural-language tasks, but adapting them to knowledge outside their static pretraining corpus remains difficult. Retrieval-augmented generation (RAG)~\citep{lewis2020retrieval} addresses this limitation by conditioning a frozen LLM on documents retrieved from an external corpus. However, retrieval alone does not solve the full adaptation problem. After relevant documents are retrieved, the model must still decide how to use them for a new task, domain, or query distribution.

Existing RAG systems usually address this problem in one of three ways. The first is to keep both the retriever and generator fixed, and simply prepend retrieved documents to the input. This strategy is efficient, but it treats retrieved documents as static evidence and gives the model no mechanism to adjust how evidence should be used. The second is to fine-tune the retriever, the generator, or both. This can improve task performance, but it requires additional training and can be expensive when the task or domain changes. The third is to use in-context learning (ICL), where a few input-output examples are provided at inference time. ICL is attractive because it avoids full model retraining, but it is still unclear whether these examples merely provide extra demonstrations, or whether they can induce a more systematic update to how a RAG model uses retrieved evidence.

This paper asks a simple question: \textit{can retrieved evidence and a few RAG examples act not only as context to read from, but also as a signal for adapting how the model uses evidence?} Answering this question requires connecting two views that have mostly been studied separately. On one side, recent theory shows that, under linear self-attention, in-context learning can implement gradient descent on the examples in the context~\citep{von2023transformers, akyurek2022what, mahankali2023one}. This suggests that context can behave like a forward-pass update, rather than only as additional input text. On the other side, RAG introduces structure that is absent from standard ICL theory: a query, retrieved documents, query-evidence interactions, and a generator that must combine them to produce an answer. It remains unclear whether the gradient-descent view of ICL extends to retrieval-augmented prediction, and whether such a view can guide practical adaptation in real RAG systems.

We study RAG from this in-context optimization perspective. Our goal is not to claim that modern retrieval-augmented LLMs literally perform gradient descent during inference. Instead, we use a controlled linear setting to identify where such an update would act. In linear RAG, the relevant update acts on the interaction between the query and retrieved evidence. This gives a simple design principle for LLM-scale RAG: rather than changing which documents are retrieved, we adapt how the frozen generator uses the retrieved documents.
%%%%
Guided by this principle, we propose a forward-only adaptation method for frozen RAG LLMs. 
The method keeps both the external retriever and the LLM backbone fixed, and adapts only a lightweight generator-side evidence-use interface implemented with LoRA.  
At inference time, given new few-shot RAG demonstrations, the predictor produces the update in a single forward pass, enabling the generator to adjust how it uses retrieved documents without re-training on the new dataset.
% Guided by this principle, we propose a forward-only adaptation method for frozen RAG LLMs. 
% The method keeps both the external retriever and the LLM backbone fixed, and adapts only a lightweight generator-side evidence-use interface implemented with LoRA. 
% A context-conditioned predictor maps a few RAG-formatted demonstrations to an update of this interface, enabling the generator to adjust how it uses retrieved documents without backpropagation at inference time.

We develop this idea in three steps. \textbf{First}, we prove that one linear self-attention layer can implement one gradient-descent step on a unified linearized RAG objective covering both projection-based and dot-product retrieval interfaces. \textbf{Second}, we test how far this correspondence extends beyond the exact construction. A trained self-attention layer closely matches the constructed gradient-descent predictor under controlled linear shifts, varying document counts, and stacked depths, while nonlinear architectures and real-world regression data reveal a clear dependence on feature distribution. \textbf{Third}, we use the optimization view to guide LLM-scale RAG adaptation. Across seven QA benchmarks, two retrievers, and two frozen LLM backbones, the predicted update improves a shared-interface baseline, transfers to held-out tasks, and approaches test-time gradient adaptation at much lower per-query cost.
Our contributions are summarized as follows:
\begin{itemize}
    \item \textbf{An in-context optimization view of linear RAG.}
    We extend the ICL-as-gradient-descent perspective from generator-only prediction to retrieval-augmented prediction. We prove that one linear self-attention layer can implement one gradient-descent step on a unified linearized RAG loss covering both linear-projection and dot-product retrieval interfaces. We also show that stacking $K$ linear self-attention layers gives a multi-step view of in-context optimization for linear RAG.

    \item \textbf{A boundary analysis beyond the exact linear setting.}
    We test when the linear construction remains predictive and when it breaks. On synthetic linear regression tasks, a trained self-attention layer closely matches the constructed gradient-descent predictor under distribution shift, varying document counts, and stacked depths. On nonlinear architectures and four real-world regression datasets, the alignment degrades in a structured way and becomes sensitive to feature distribution.

    \item \textbf{Adapting evidence use without test-time backpropagation.}
    We use the optimization view to guide adaptation in frozen RAG LLMs. Rather than changing the external retriever, we adapt a generator-side evidence-use interface implemented with Q/K/V LoRA modules. A small context-conditioned predictor amortizes the autograd-defined $K$-step update to this interface. Across seven QA benchmarks, two backbones, and two retrievers, the predicted update improves a shared-interface baseline, transfers to held-out domains, and approaches test-time gradient adaptation at much lower per-query cost.
\end{itemize}
\section{Related Work}\label{sec:related}

Retrieval-augmented generation (RAG) conditions a language model on documents retrieved from an external corpus~\citep{lewis2020retrieval, guu2020retrieval, karpukhin2020dense, izacard2021leveraging, borgeaud2022retro, zhang2025survey}. Prior work explores retrieval, prompting, evidence fusion, memory-based knowledge access, and learning-based adaptation~\citep{ram2023incontext, asai2024self, huang2024ritek, zhang2025training, li2026ricepoturningretrievalinteractions, li2026efficienteffectiveinternalmemory}. Our focus is complementary: rather than changing which documents are retrieved, we study how a frozen generator can adapt its use of already-retrieved evidence. We position this contribution relative to three lines of work.

\paragraph{In-context learning as gradient descent.}
A growing line of theory interprets in-context learning as implicit optimization. Under linear self-attention, a single ICL forward pass can implement one gradient-descent step, and stacked layers can implement multiple steps~\citep{von2023transformers, akyurek2022what, mahankali2023one, dai2023gpt}. Later work extends this view to preconditioned gradient descent~\citep{ahn2023transformers}, in-context algorithm selection~\citep{bai2023transformers}, the role of depth~\citep{vladymyrov2024linear, gatmiry2024can}, and kernel-regression interpretations of attention~\citep{shen2025understanding, ren2024towards}. These analyses mainly study generator-only settings, often through linear regression or simplified attention. RAG introduces additional structure, including retrieved documents, query-evidence interactions, and evidence-conditioned generation. We extend the gradient-descent view to a linearized RAG setting and use it to identify where evidence-use adaptation should act.

\paragraph{Context-conditioned weight prediction.}
Another line of work learns auxiliary networks that produce model updates from a small context. HyperTuning~\citep{phang2023hypertuning} predicts soft prompts or low-rank weights from few-shot examples. HyperFlow~\citep{kim2025hyperflow} learns support-conditioned fine-tuning dynamics. MAC~\citep{tack2024online} maps documents into memory modulations. MEND~\citep{mitchell2022mend} maps fine-tuning gradients into knowledge edits. \textsc{RAG-GD} follows the broad template of predicting an update from context, but differs in both the target and the adaptation site. The target is not a downstream task loss, a meta-learning objective, a memory objective, or a single editing gradient. Instead, the predictor matches an autograd-defined $K$-step SGD update induced by RAG-formatted demonstrations. The adapted parameters are also restricted to a generator-side evidence-use interface, while the retriever and backbone remain fixed.

\paragraph{Test-time adaptation.}
Standard adaptation either updates model parameters before deployment, as in fine-tuning and LoRA~\citep{hu2022lora}, or leaves the model unchanged at inference, as in pure ICL. Test-time training~\citep{sun2020testtime} lies between these extremes by updating parameters for each test instance, but this requires per-instance backpropagation and becomes expensive for large LLMs. Recent studies examine system-level strategies spanning ICL, fine-tuning, trainable RAG, and adaptive test-time compute~\citep{wang2024benchmarking, mosbach2023few, li2024benchmarking, li2026signaloppositemeaningdirectioninformed}. \textsc{RAG-GD} targets the same goal of adapting at inference time, but amortizes the update: a small predictor emits a LoRA update to the generator's evidence-use interface in one forward pass. Thus, it avoids backpropagation through the LLM at deployment while keeping both the external retriever and the frozen backbone unchanged.
\section{A Linear RAG Setting Where Self-Attention Implements Gradient Descent}
\label{sec:method}

This section establishes the linear-regime basis for our in-context optimization view of RAG.
We study a controlled setting in which retrieval-augmented prediction and gradient descent can be connected exactly.
The goal is not to model modern RAG systems literally: real retrievers involve discrete document selection, and modern LLMs are deep and nonlinear.
Instead, we isolate a differentiable retrieval-augmented prediction problem and show that one linear self-attention layer can realize the prediction shift produced by one gradient-descent step.
The proof and explicit construction are in Appendix~\ref{Appendix: Proposition1}, and the derivations for the retrieval variants are in Appendix~\ref{appendix:linearRAG}.

\subsection{Self-Attention}
\label{sec:motivation}

We begin with a multi-head self-attention block parameterized by 
$\theta = \{P_h, W_{h,V}, W_{h,K}, W_{h,Q}\}_{h=1}^H$. 
Given tokens $\{e_1,\ldots,e_N\} \subset \mathbb{R}^d$, the update for token $e_j$ is
\begin{equation}
e_j 
\gets 
e_j + \mathrm{SA}_\theta(j, \{e_i\}_{i=1}^N)
=
e_j + \sum_h P_h V_h \,\mathrm{softmax}(K_h^\top q_{h,j}),
\label{eq:sa_update}
\end{equation}
where $V_h$, $K_h$, and $q_{h,j}$ are the value matrix, key matrix, and query vector for head $h$.
Following~\citep{von2023transformers, vladymyrov2024linear}, we remove the softmax and bias terms to obtain the linear self-attention (LSA) update:
\begin{equation}
e_j 
\gets 
e_j + \mathrm{LSA}_\theta(j, \{e_i\}_{i=1}^N)
=
e_j + \sum_h P_h V_h K_h^\top q_{h,j}.
\label{eq:lsa_update}
\end{equation}

\subsection{A Unified Linearized RAG Predictor}

We use a linearized abstraction of retrieval-augmented prediction.
Rather than modeling discrete top-$k$ selection, this abstraction captures a differentiable interface in which query features and retrieval-derived features jointly determine the prediction.
Both a projection-based retrieval interface~\citep{lewis2020retrieval} and a dot-product retrieval interface~\citep{karpukhin2020dense} can be written as
$y = W_1 x_1 + W_2 x_2$,
where $x_1$ denotes the query-side feature and $x_2$ denotes the retrieval-derived feature.
%%%%%
For the projection-based interface, we set $x_1 = x_q$, $x_2 = D$, and $W_2 \triangleq W_1 W_d$, where $W_d$ projects document embeddings into the prediction space.
For the dot-product interface, we set $x_1 = x_2 = x_q$ and
$
W_2 = W_z\left(\sum_i d_i d_i^\top\right)M^\top,
$
where $M$ parameterizes query-document similarity.
For tractability, we use the shared-encoder simplification $M = W_e^\top W_e$, so $M$ is symmetric.
The general DPR formulation~\citep{karpukhin2020dense} allows separate query and document encoders, with $M = W_q^\top W_d$.
Full derivations are provided in Appendix~\ref{appendix:linearRAG}.

\subsection{Optimization Objective}

Given training examples $\{(x_1^i, x_2^i, y_i)\}_{i=1}^N$, we consider the squared loss
\begin{equation}
L(W_1, W_2)
=
\frac{1}{2N}
\sum_{i=1}^N
\left\|
W_1 x_1^i + W_2 x_2^i - y_i
\right\|^2.
\label{eq:rag_loss}
\end{equation}
One gradient-descent step with learning rate $\eta$ gives
\begin{equation}
\Delta W_k
=
-\eta \nabla_{W_k} L
=
-\frac{\eta}{N}
\sum_{i=1}^N
\left(
W_1 x_1^i + W_2 x_2^i - y_i
\right)
(x_k^i)^\top,
\qquad k \in \{1,2\}.
\label{eq:gd_step}
\end{equation}
For a query token with features $(x_1,x_2)$, the corresponding prediction shift is
$
\Delta y
\triangleq
\Delta W_1 x_1 + \Delta W_2 x_2.
$
Thus, $\Delta y$ is the change in prediction after updating $W_k$ to $W_k' = W_k + \Delta W_k$.

\subsection{Linear Self-Attention Reproduces One Gradient Step}

\begin{lemma}[Linear self-attention implements one RAG gradient step]
\label{lemma:equivalence}
Consider a 1-head linear self-attention layer, context tokens $e_i = (x_1^i, x_2^i, y^i)$ for $i = 1,\ldots,N$, and a query token $e_j = (x_1^j, x_2^j, y^j)$.
Let $\Delta W_1$ and $\Delta W_2$ be the one-step gradient-descent updates in Eq.~\ref{eq:gd_step}.
There exist matrices $W_K, W_Q, W_V$ and an output projection $P$ such that one LSA update changes only the $y$-coordinate of $e_j$:
\begin{equation}
e_j
\leftarrow
e_j
+
\left(
0,\,
0,\,
\Delta W_1 x_1^j + \Delta W_2 x_2^j
\right).
\label{eq:lemma_update}
\end{equation}
Equivalently,
\begin{equation}
P V K^\top q_j
=
\left(
0,\,
0,\,
\Delta W_1 x_1^j + \Delta W_2 x_2^j
\right).
\end{equation}
Therefore, the LSA update exactly matches the prediction shift induced by one gradient-descent step on the unified linearized RAG predictor.
\end{lemma}
%%%%
The construction is given in Appendix~\ref{Appendix: Proposition1}.
Intuitively, the value projection encodes the residual $W_1x_1^i + W_2x_2^i - y^i$.
The key-query interaction computes the inner products $(x_1^i)^\top x_1^j$ and $(x_2^i)^\top x_2^j$.
The output projection then writes the resulting weighted residual sum into the query token's prediction coordinate.

This construction also gives a controlled multi-step analogue.
If each LSA layer represents one gradient-like update, then after $K$ layers,
\begin{equation}
\hat{y}^{(K)}_{N+1}
=
\hat{y}^{(0)}_{N+1}
+
\sum_{t=0}^{K-1}
\left(
\Delta W^{(t)}_1 x^1_{N+1}
+
\Delta W^{(t)}_2 x^2_{N+1}
\right),
\label{eq:stacked}
\end{equation}
where $\Delta W^{(t)}_1$ and $\Delta W^{(t)}_2$ are the implicit updates represented by layer $t$.
We use this multi-step view as a linear-regime guide rather than as a literal claim about frozen LLM computation.
In later sections, this view motivates a forward-only mechanism that adapts how a frozen generator uses retrieved evidence.

\section{Testing the Boundary of the Linear Correspondence}
\label{sec:empirical_bridge}
Lemma~\ref{lemma:equivalence} gives an exact correspondence in a controlled linear setting.
We now ask how far this correspondence remains predictive when the setting is varied.
The experiments have two goals: first, to verify that a trained linear self-attention layer can reproduce the constructed gradient-descent predictor; second, to identify where the correspondence begins to break beyond the exact regime.
\subsection{Linear-Regime Verification}
\label{sec:linear_emp}
Each token concatenates an input feature, a retrieval-derived feature, and a target, $e_i=(x_i,z_i,y_i)$ for $i=1,\ldots,N$.
The auxiliary slot $z_i$ instantiates the unified RAG view.
For the projection-based interface, $z_i$ is a document-derived feature.
For the dot-product interface, $z_i=x_i$, and document information is injected into the keys and values.
%%%%%
We train an LSA layer $\theta$ to minimize expected squared error across tasks, using minibatch SGD over freshly sampled tasks.
Following prior work~\citep{garg2022can, von2023transformers}, each task is generated from a teacher with weights $W_\tau \sim \mathcal{N}(0,I)$.
Inputs are sampled as $x_{\tau,i}\sim\mathcal{U}(-1,1)^{n_I}$, and targets are generated by
$y_{\tau,i}=W_\tau^1x_{\tau,i}^1+W_\tau^2x_{\tau,i}^2$.
We set $N=n_I=10$ and sweep the document count $k \in \{2,5,10,25\}$.
%%%%%%%%
We compare the trained layer $\theta^*$ with the constructed predictor that exactly realizes one gradient-descent step on the unified RAG loss.
On $T_{\mathrm{val}}=10^4$ held-out tasks, we report the prediction difference
$\|\hat{y}_{\theta^*}-\hat{y}_{\theta,\mathrm{rag}}\|_2$,
the cosine similarity between input sensitivities $\partial \hat{y}/\partial x_{\mathrm{test}}$,
and the corresponding sensitivity $\ell_2$ difference.
Details are in Appendix~\ref{appendix:linear}.
%%%%%%%%%
\begin{figure*}[t]
    \centering
    \begin{minipage}{0.23\textwidth}
        \includegraphics[width=\linewidth]{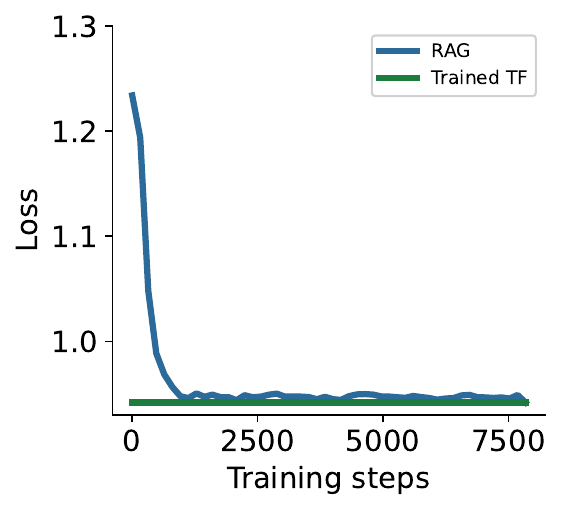}
    \end{minipage}
    \begin{minipage}{0.25\textwidth}
        \includegraphics[width=\linewidth]{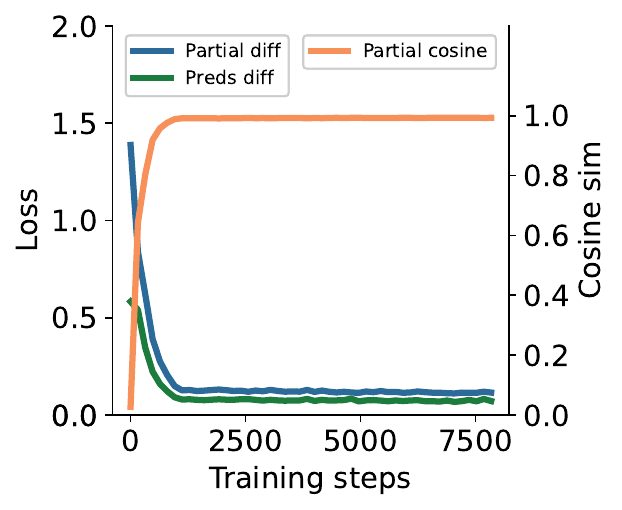}
    \end{minipage}
    \begin{minipage}{0.23\textwidth}
        \includegraphics[width=\linewidth]{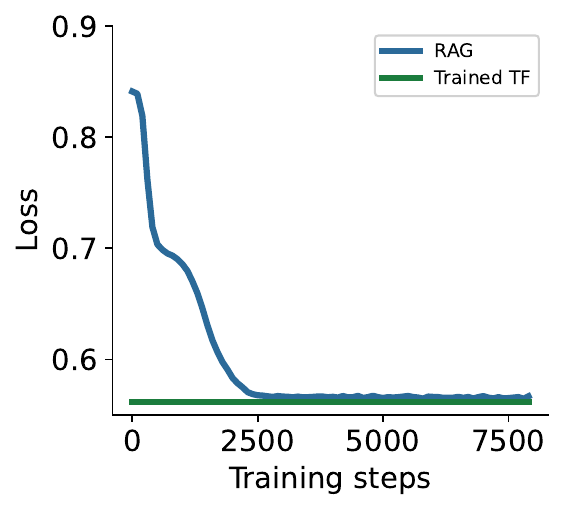}
    \end{minipage}
    \begin{minipage}{0.25\textwidth}
        \includegraphics[width=\linewidth]{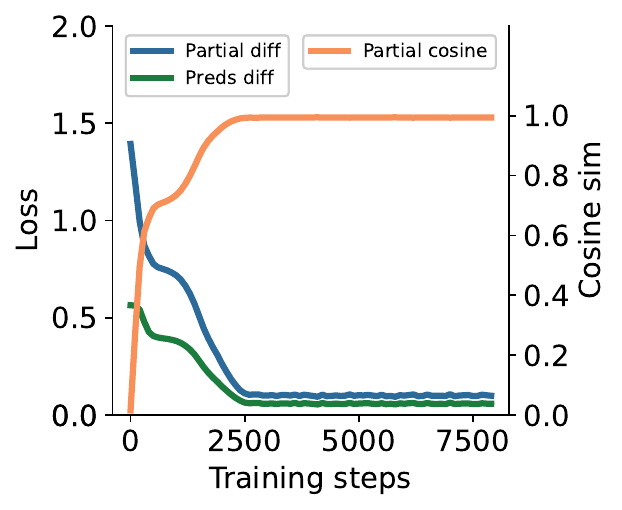}
    \end{minipage}
    \caption{
    Single-layer LSA reproduces one gradient-descent step on the unified linearized RAG loss.
    The two left panels report the projection-based interface, and the two right panels report the dot-product interface.
    Across both variants, the trained LSA layer and the constructed gradient-descent predictor are nearly indistinguishable on held-out tasks.
    }
    \label{fig:layer1_results}
    \vspace{-10pt}
\end{figure*}
Figure~\ref{fig:layer1_results} verifies the construction for both retrieval interfaces.
The trained LSA layer closely matches the constructed predictor: the loss difference is small, the sensitivity cosine is close to $1$, and the sensitivity $\ell_2$ difference is negligible.
This numerically confirms the algebraic correspondence in Lemma~\ref{lemma:equivalence}.

\subsection{Controlled Stress Tests}
\label{sec:linear_stress}

We next test whether the agreement persists under controlled changes within the linear regime.
We vary the document count, shift the test-input distribution, and stack LSA layers with shared parameters.
When sweeping $n \in \{2,5,10,25\}$ and shifting the test-input range to $\alpha \in \{0.5,1,1.5,2\}$ while keeping training fixed at $\alpha=1$, the loss difference between the trained Transformer and the gradient predictor remains small (Figure~\ref{figure:differ_doc_distrubution}, Appendix~\ref{appendix:linear}).
%%%%%%%
Stacking LSA layers further supports the multi-step picture in Eq.~\ref{eq:stacked}.
At depths $2$ and $5$, the loss and prediction differences remain small across document counts.
The residual gap at $\mathrm{Docs}=25$ also shrinks as depth increases (Figure~\ref{figure:layer_different_doc_R2}, Appendix~\ref{appendix:linear}).
The projection-based interface shows similar behavior (Appendix~\ref{section:RAG_non_linear_layers}).
These results suggest that the linear correspondence is not a fragile single-step artifact, but remains stable under controlled linear extensions.

\subsection{Nonlinear Stress Test}
\label{sec:nonlinear_emp}

We then examine where the correspondence begins to break.
We add MLP layers after the input embedding and evaluate on four real-world regression datasets: California Housing, Bike Sharing, Wine Quality, and Predict Calorie Expenditure.
We focus on the dot-product interface throughout.
Dataset details are in Appendix~\ref{con:dataset_detals}.
%%%%%%%
This experiment is diagnostic.
We do not claim that normalization solves RAG adaptation.
Instead, we use normalization to control the feature geometry that interacts with dot-product retrieval.
We compare Z-score~\citep{bishop2006pattern}, Min--Max~\citep{bishop2006pattern}, rank-based normalization~\citep{conover1999practical}, and Tanh normalization.
The training set is used as the retrieval corpus and is normalized with Z-score throughout.
Only the input-side normalization is varied, and alignment is measured using the same metrics as in Section~\ref{sec:linear_emp}.
%%%%%%%%5
\begin{figure*}[t]
    \centering
    \includegraphics[width=0.24\textwidth]{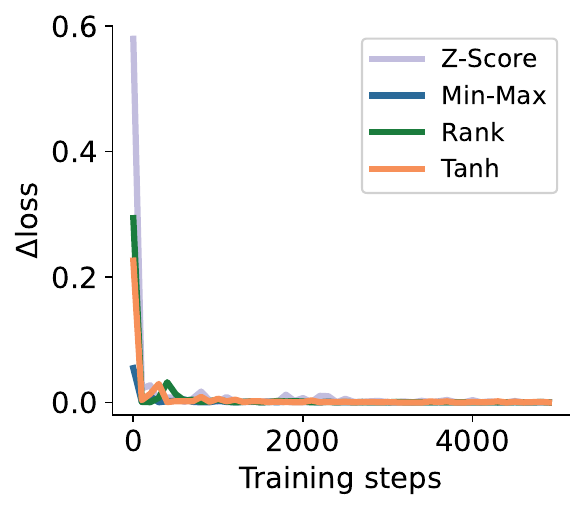}%
    \includegraphics[width=0.24\textwidth]{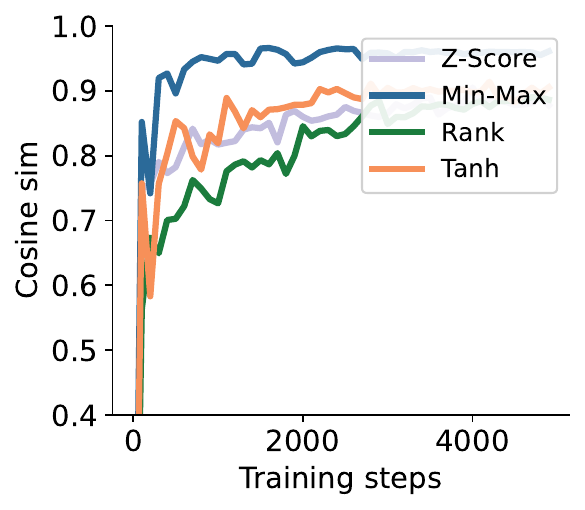}%
    \includegraphics[width=0.24\textwidth]{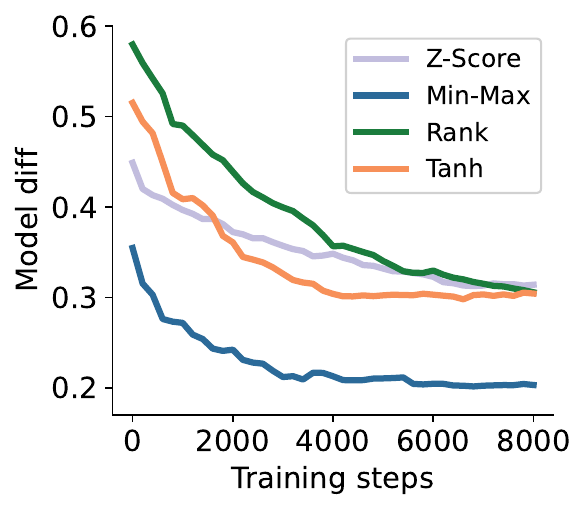}%
    \includegraphics[width=0.24\textwidth]{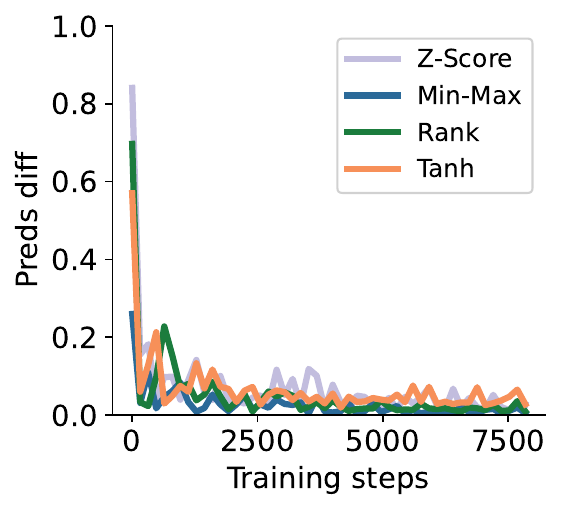}\\[1ex]
    \includegraphics[width=0.24\textwidth]{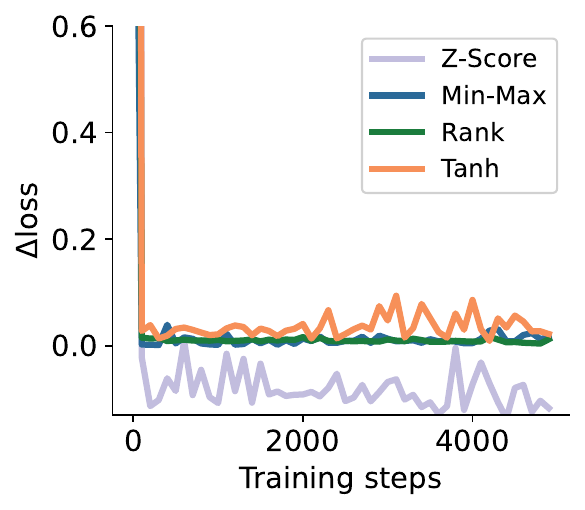}%
    \includegraphics[width=0.24\textwidth]{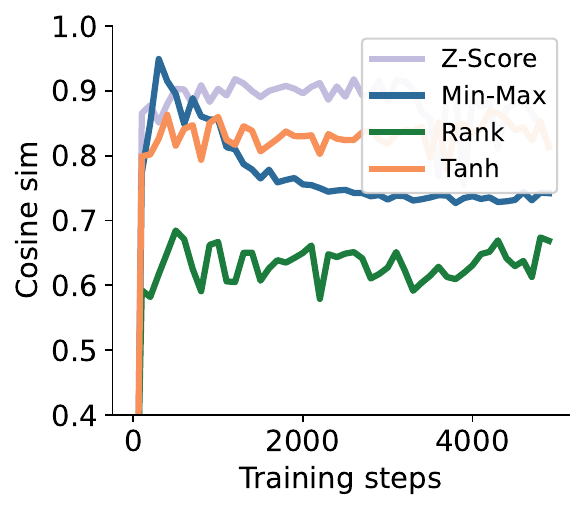}%
    \includegraphics[width=0.24\textwidth]{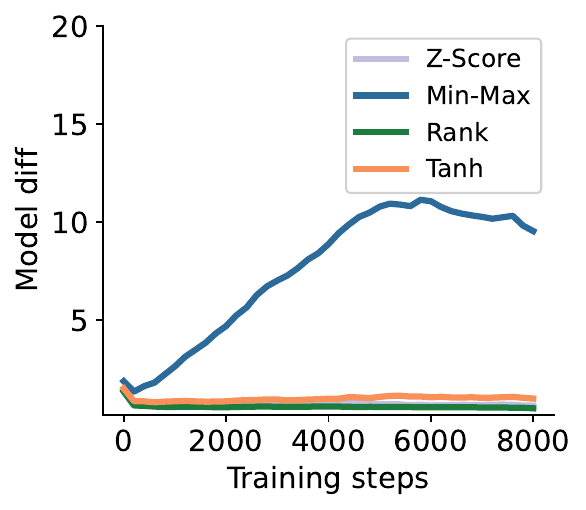}%
    \includegraphics[width=0.24\textwidth]{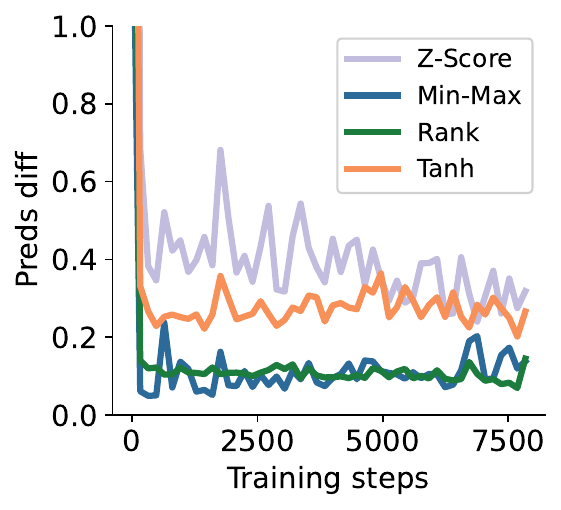}%
    \caption{
    Effect of input normalization on the alignment between the trained nonlinear Transformer and the gradient-descent predictor under the dot-product interface.
    \textit{Top row:} Bike Sharing.
    \textit{Bottom row:} California Housing.
    Columns report loss difference, sensitivity cosine, model difference, and prediction difference.
    Min--Max normalization closely matches the gradient-descent predictor on Bike Sharing, where features are bounded and roughly uniform.
    On California Housing, where features are skewed and heavy-tailed, the alignment degrades.
    }
    \label{figure:real_data_bike_c}
    \vspace{-15pt}
\end{figure*}
%%%%
Figure~\ref{figure:real_data_bike_c} shows two representative cases.
On Bike Sharing, Min--Max normalization gives the closest agreement between the trained nonlinear Transformer and the gradient-descent predictor, likely because the features are bounded and not dominated by outliers.
On California Housing, the alignment is weaker: skewed and heavy-tailed features make dot-product geometry more sensitive to outliers.
The sensitivity cosine drops, the model difference grows, and prediction differences become less stable.
%%%%%%
The same pattern appears on the remaining datasets.
Predict Calorie Expenditure behaves similarly to Bike Sharing, while Wine Quality behaves more like California Housing (Figure~\ref{figure:cals_win}, Appendix~\ref{appendix:normalization}).
Overall, the linear optimization view remains informative when feature geometry is stable, but becomes less predictive when retrieval-derived dot products are dominated by skewed or heavy-tailed features.
This empirical boundary supports our use of the linear construction as a guide for adaptation, rather than as a literal model of LLM computation.
Next, we use this view to design a forward-only update to the generator-side evidence-use interface in LLM-scale RAG.

\section{LLM-Scale RAG: Amortizing the Gradient-Descent Update}
\label{sec:llm_probe}
We now instantiate this view as \textsc{RAG-GD}, a forward-only adaptation method for frozen billion-parameter RAG LLMs. In this setting, we do not assume an exact equivalence between an LLM forward pass and gradient descent. Instead, we use gradient descent as an operational target for adapting how the generator uses retrieval-conditioned information.
Given a few RAG-formatted demonstrations, we use autograd during training to compute the update that gradient descent would make to the generator-side retrieval interface.
We then train a lightweight predictor to approximate this update. At inference time, the predictor produces the update in a single forward pass, without further training the RAG system or backpropagating through the frozen LLM.

Concretely, we first train a base retrieval adapter $W_0^{\mathrm{ret}}$ on RAG-formatted examples from the NQ training split.
It is a low-rank LoRA perturbation to the Q/K/V projections of every attention layer in a frozen LLM. All inputs are RAG-formatted instances $(x,\mathcal{D}_x,y)$, where $y$ is the gold answer. 
In this work, we use a fixed external retriever, either BM25 or E5, to select the retrieved documents $\mathcal{D}_x$ for each question $x$ from a fixed corpus.
%%%%%%
This adapter serves as a generator-side retrieval interface: it does not select documents, but modulates how the generator uses evidence selected by BM25 or E5. 
The predictor $g_\phi$ is then meta-trained on few-shot support contexts $C=\{(x_i,\mathcal{D}_i,y_i)\}_{i=1}^N$ from NQ, TriviaQA, HotpotQA, 2WikiMultiHopQA, and MuSiQue. 
PopQA and Bamboogle are held out from both stages and used only for evaluation.

\subsection{Supervision Target: Autograd-Defined Interface Update}
\label{sec:llm_target}

The supervision target is the update that $K$ SGD steps would produce on the generator-side retrieval interface using a support context $C$. 
Starting from $W_{\mathrm{ret}}^{(0)}=W_0^{\mathrm{ret}}$, we run
\begin{equation}
\Delta W_{\mathrm{GD}}^{(K)}(C)
=
W_{\mathrm{ret}}^{(K)} - W_0^{\mathrm{ret}},
\qquad
W_{\mathrm{ret}}^{(t+1)}
=
W_{\mathrm{ret}}^{(t)}
-
\eta \nabla \mathcal{L}(W_{\mathrm{ret}}^{(t)}; C),
\label{eq:gd_target}
\end{equation}
for $t=0,\ldots,K-1$, where $\mathcal{L}$ is the answer-token cross-entropy conditioned on the question and retrieved documents. 
We compute $\Delta W_{\mathrm{GD}}^{(K)}(C)$ with autograd and detach it from the predictor optimizer.
%%%%%%
The practical method is amortized test-time adaptation through a context-conditioned LoRA predictor, rather than a direct implementation of Lemma~\ref{lemma:equivalence}.
The linear analysis motivates two design decisions: \emph{where to adapt}, since the support-induced update acts on the query--evidence interaction, and \emph{what to predict}, since this update is a gradient-descent step in the linear setting.
We therefore adapt the generator-side evidence-use interface and train the predictor to match the $K$-step gradient update rather than a downstream task loss.

\subsection{Predictor Architecture and Matching Objective}
\label{sec:llm_predictor}

The predictor $g_\phi$ is a context encoder with per-layer, per-projection update heads. 
Each demonstration $(x_i,\mathcal{D}_i,y_i)\in C$ is formatted by concatenating the question, retrieved documents, and gold answer. 
The frozen LLM with the base adapter encodes each sequence, and we use the EOS hidden state $h_i\in\mathbb{R}^{d_h}$. 
We aggregate demonstrations by mean pooling, $\bar h(C)=\frac{1}{N}\sum_{i=1}^{N}h_i$.
For each layer $\ell$ and projection type $\pi\in\{Q,K,V\}$, the update head outputs
$
\widetilde{\Delta W}_{\ell,\pi}
=
U_{\ell,\pi}V_{\ell,\pi}^{\top}
\in \mathbb{R}^{d\times d},
$
where $U_{\ell,\pi},V_{\ell,\pi}\in\mathbb{R}^{d\times r}$ are generated by a two-layer MLP from $\bar h(C)$. 
The rank $r$ matches the base adapter, so $g_\phi(C)$ has the same shape as $W_0^{\mathrm{ret}}$.
%%%
We train $g_\phi$ to match the autograd-defined target per layer and projection type. 
Let
$
\Delta W_{\ell,\pi}^{\star}
\triangleq
[\Delta W_{\mathrm{GD}}^{(K)}(C)]_{\ell,\pi}.
$
The matching loss is
\begin{equation}
\mathcal{L}_{\mathrm{match}}(\phi;C)
=
\sum_{\ell,\pi}
\left[
1-
\frac{
\langle \widetilde{\Delta W}_{\ell,\pi},\Delta W_{\ell,\pi}^{\star}\rangle
}{
\|\widetilde{\Delta W}_{\ell,\pi}\|_F
\|\Delta W_{\ell,\pi}^{\star}\|_F
}
+
\lambda
\left|
\log
\frac{
\|\widetilde{\Delta W}_{\ell,\pi}\|_F
}{
\|\Delta W_{\ell,\pi}^{\star}\|_F
}
\right|
\right],
\label{eq:matching}
\end{equation}
where the cosine term matches direction and the log-magnitude term matches scale. 
We use $\lambda=0.1$ throughout.
At deployment, the predictor emits $\widetilde{\Delta W}(C)$ in one forward pass, and the frozen LLM answers with the adapted interface $W_0^{\mathrm{ret}}+\widetilde{\Delta W}(C)$.

\subsection{Benchmarks, Baselines, and Metrics}
\label{sec:exp-setup}
\paragraph{Benchmarks.} We evaluate on seven open-domain QA benchmarks: NQ, TriviaQA, PopQA, HotpotQA, 2WikiMultiHopQA, MuSiQue, and Bamboogle.
NQ, TriviaQA, and PopQA are single-hop, while the remaining four are multi-hop. 
We use Qwen 2.5-7B-Instruct~\citep{yang2024qwen} and Llama 3.1-8B-Instruct~\citep{dubey2024llama} as frozen backbones. 
Each query is augmented with the top five documents from BM25 or E5-large~\citep{wang2022text}, using the same retrieval cache for all methods. 
The support size is $N=3$, and we train predictors with $K\in\{1,5,10\}$.

\paragraph{Baselines and metrics.} We compare \textsc{RAG-GD} with six baselines.
\textbf{Query Only} uses no retrieved documents. 
\textbf{Vanilla RAG} prepends retrieved documents to the prompt. 
\textbf{Base adapter} applies $W_0^{\mathrm{ret}}$ without context-conditioned perturbation. 
\textbf{+ few-shot} variants concatenate support demonstrations into Vanilla RAG or Base adapter prompts.
\textbf{Prompt Tuning}~\citep{lester2021power} learns a soft prefix using the same supervision pool as $W_0^{\mathrm{ret}}$.
\textbf{HyperTuning}~\citep{phang2023hypertuning} trains a hypernetwork to generate task-specific adaptation parameters from few-shot examples by optimizing the prediction loss of a frozen downstream model.
\textbf{TT-SGD} performs $K$ SGD steps on $C$ at test time and serves as the non-amortized reference.
\textsc{RAG-GD} shares $W_0^{\mathrm{ret}}$ with Base adapter and adds only $\widetilde{\Delta W}(C)$. 
Table~\ref{tab:main} reports the headline comparison against no-perturbation baselines. 
Full results for Prompt Tuning, HyperTuning, TT-SGD, and + few-shot variants are in Appendix~\ref{appendix:full_qa_table}.
We report SQuAD-style~\citep{rajpurkar2016squad} exact match (EM) and token-overlap F1.

% --- Table 1 ---

\begin{table*}[t]
\centering
% \scriptsize
% \setlength{\tabcolsep}{2.8pt}
\caption{
Main QA results across seven benchmarks, two frozen LLM backbones, and two retrievers. 
\textsc{RAG-GD} ($K{=}5$) applies a context-conditioned update on top of the same static retrieval adapter $W_0^{\mathrm{ret}}$ used by Base adapter. 
Bold values mark the best result per column for each backbone--retriever pair.
Full context-conditioned baseline results are in Appendix~\ref{appendix:full_qa_table}.
}
\renewcommand{\arraystretch}{1}
\resizebox{\textwidth}{!}{
\begin{tabular}{llcccccccccccccccc}
\toprule
\multirow{3}{*}{\textbf{Method}} & \multirow{3}{*}{\textbf{Retriever}}
& \multicolumn{6}{c}{\textbf{Single-Hop QA}}
& \multicolumn{8}{c}{\textbf{Multi-Hop QA}}
& \multicolumn{2}{c}{\textbf{Avg.}} \\
\cmidrule(lr){3-8} \cmidrule(lr){9-16} \cmidrule(lr){17-18}
& & \multicolumn{2}{c}{\textbf{NQ}} & \multicolumn{2}{c}{\textbf{TriviaQA}} & \multicolumn{2}{c}{\textbf{PopQA}}
& \multicolumn{2}{c}{\textbf{HotpotQA}} & \multicolumn{2}{c}{\textbf{2Wiki}} & \multicolumn{2}{c}{\textbf{MuSiQue}} & \multicolumn{2}{c}{\textbf{Bamboogle}}
& & \\
\cmidrule(lr){3-4} \cmidrule(lr){5-6} \cmidrule(lr){7-8} \cmidrule(lr){9-10} \cmidrule(lr){11-12} \cmidrule(lr){13-14} \cmidrule(lr){15-16}
& & EM & F1 & EM & F1 & EM & F1 & EM & F1 & EM & F1 & EM & F1 & EM & F1 & EM & F1 \\
\midrule
% \rowcolor{gray!20}
\multicolumn{18}{c}{\textbf{Qwen 2.5-7B}} \\
\midrule
Query Only & --
& 15.95 & 24.28 & 43.33 & 49.51 & 16.02 & 19.76 & 18.40 & 25.39 & 23.91 & 28.12 & 3.80 & 10.57 & 11.20 & 18.02 & 18.94 & 25.09 \\
\midrule
\multirow{2}{*}{Vanilla RAG}
& BM25 & 27.61 & 36.66 & 58.24 & 65.77 & 28.84 & 33.18 & 31.28 & 41.25 & 27.87 & 33.24 & 5.87 & 13.05 & 10.40 & 21.16 & 27.16 & 34.90 \\
& E5   & 39.16 & 50.03 & 62.99 & 70.80 & 44.03 & 50.21 & 32.45 & 42.21 & 25.48 & 31.43 & 5.79 & 12.77 & 18.40 & 26.74 & 32.61 & 40.60 \\
\midrule
\multirow{2}{*}{Base adapter}
& BM25 & 32.57 & 41.45 & 60.11 & 67.93 & 31.55 & 35.63 & 32.31 & 43.45 & 28.22 & 34.13 & 6.41 & 15.46 & 16.80 & 26.01 & 29.71 & 37.72 \\
& E5   & 41.77 & 51.22 & 63.31 & 71.62 & 47.05 & 51.82 & 33.78 & 44.64 & 27.89 & 33.97 & 6.95 & 15.76 & 18.40 & 28.78 & 34.16 & 42.54 \\
\midrule
\rowcolor{ragtableblue!12}
& BM25 & \textbf{34.46} & \textbf{43.54} & \textbf{63.27} & \textbf{70.69} & \textbf{33.22} & \textbf{37.69} & \textbf{35.54} & \textbf{47.14} & \textbf{28.86} & \textbf{34.48} & \textbf{9.26} & \textbf{19.73} & \textbf{22.40} & \textbf{32.85} & \textbf{32.43} & \textbf{40.87} \\

\rowcolor{ragtableblue!12}
\multirow{-2}{*}{\textcolor{ragtabletext}{\textbf{\textsc{RAG-GD} ($K{=}5$)}}}
& E5   & \textbf{42.91} & \textbf{52.71} & \textbf{65.98} & \textbf{73.60} & \textbf{48.12} & \textbf{52.61} & \textbf{35.54} & \textbf{47.00} & \textbf{29.67} & \textbf{35.47} & \textbf{9.14} & \textbf{19.26} & \textbf{25.60} & \textbf{35.12} & \textbf{36.71} & \textbf{45.11} \\
\midrule
% \rowcolor{gray!20}
\multicolumn{18}{c}{\textbf{Llama 3.1-8B}} \\
\midrule
Query Only & --
& 22.46 & 32.51 & 52.67 & 59.79 & 20.63 & 25.09 & 18.31 & 25.71 & 26.39 & 31.04 & 3.81 & 9.51 & 6.40 & 12.88 & 21.52 & 28.08 \\
\midrule
\multirow{2}{*}{Vanilla RAG}
& BM25 & 31.41 & 40.95 & 60.43 & 68.35 & 31.08 & 35.43 & 31.92 & 42.46 & 26.07 & 31.82 & 5.75 & 12.44 & 14.40 & 22.93 & 28.72 & 36.34 \\
& E5   & 40.72 & 52.23 & 64.42 & 72.58 & 45.85 & 51.48 & 32.78 & 43.19 & 23.44 & 29.46 & 6.04 & 12.25 & 24.80 & 32.12 & 34.01 & 41.90 \\
\midrule
\multirow{2}{*}{Base adapter}
& BM25 & 38.47 & 49.42 & 62.66 & 72.46 & 37.80 & \textbf{42.29} & 37.35 & 50.41 & 33.66 & 39.55 & 11.46 & 21.98 & \textbf{29.60} & \textbf{41.29} & 35.86 & 45.34 \\
& E5   & 43.46 & 54.60 & 63.90 & 74.05 & 51.69 & 55.74 & 37.31 & 49.90 & 33.45 & 39.45 & 11.83 & 22.07 & 30.40 & 40.48 & 38.86 & 48.04 \\
\midrule
\rowcolor{ragtableblue!12}
& BM25 & \textbf{40.22} & \textbf{50.01} & \textbf{66.13} & \textbf{74.28} & \textbf{37.81} & 42.19 & \textbf{38.99} & \textbf{51.14} & \textbf{34.15} & \textbf{40.04} & \textbf{12.54} & \textbf{22.61} & 28.00 & 39.61 & \textbf{36.83} & \textbf{45.70} \\
\rowcolor{ragtableblue!12}
\multirow{-2}{*}{\textcolor{ragtabletext}{\textbf{\textsc{RAG-GD} ($K{=}5$)}}}
& E5   & \textbf{45.68} & \textbf{55.84} & \textbf{67.66} & \textbf{76.07} & \textbf{52.31} & \textbf{56.48} & \textbf{39.01} & \textbf{50.94} & \textbf{33.94} & \textbf{39.83} & \textbf{13.20} & \textbf{23.47} & \textbf{32.00} & \textbf{42.08} & \textbf{40.54} & \textbf{49.24} \\
\bottomrule
\end{tabular}
}
\vspace{4pt}
\label{tab:main}
\end{table*}

\subsection{Results}
\label{sec:exp-results}

\begin{figure}[t]
  \centering
  \begin{subfigure}[b]{0.33\linewidth}
    \centering
    \includegraphics[width=\linewidth]{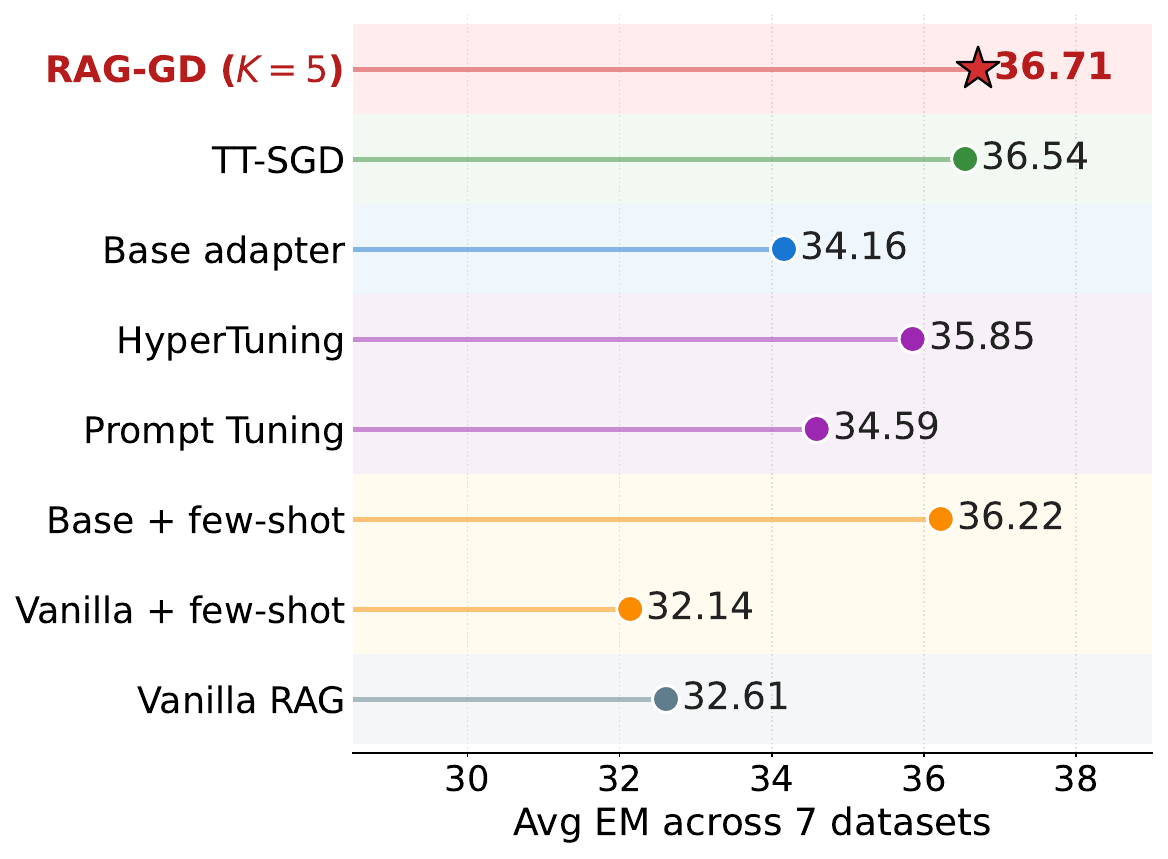}
    \caption{Method-family comparison.}
    \label{fig:method_family}
  \end{subfigure}\hfill
  \begin{subfigure}[b]{0.65\linewidth}
    \centering
    \includegraphics[width=\linewidth]{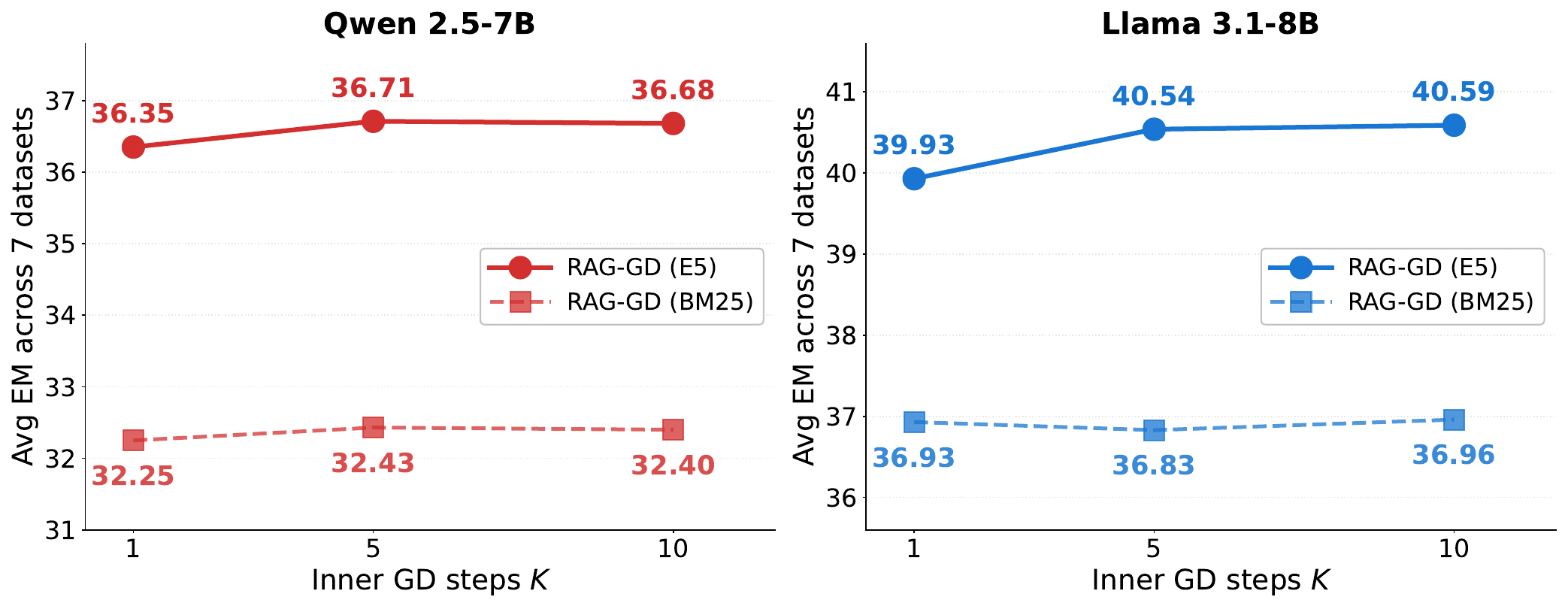}
    \caption{Robustness to inner GD depth $K$.}
    \label{fig:k_trend}
  \end{subfigure}
  \caption{
  \textbf{The gradient-derived update improves context-conditioned adaptation and is relatively insensitive to $K$.}
  \textbf{(a)} Average EM across seven QA benchmarks on Qwen 2.5-7B with E5 retrieval. 
  \textsc{RAG-GD} matches the TT-SGD reference using one predictor forward pass.
  \textbf{(b)} Average EM across $K\in\{1,5,10\}$ on both backbones and retrievers. 
  Solid lines use E5, and dashed lines use BM25.
  }
  \label{fig:family_ktrend}
  \vspace{-10pt}
\end{figure}

\begin{wrapfigure}{r}{0.45\linewidth}
  \vspace{-1.0em}
  \centering
  \includegraphics[width=\linewidth]{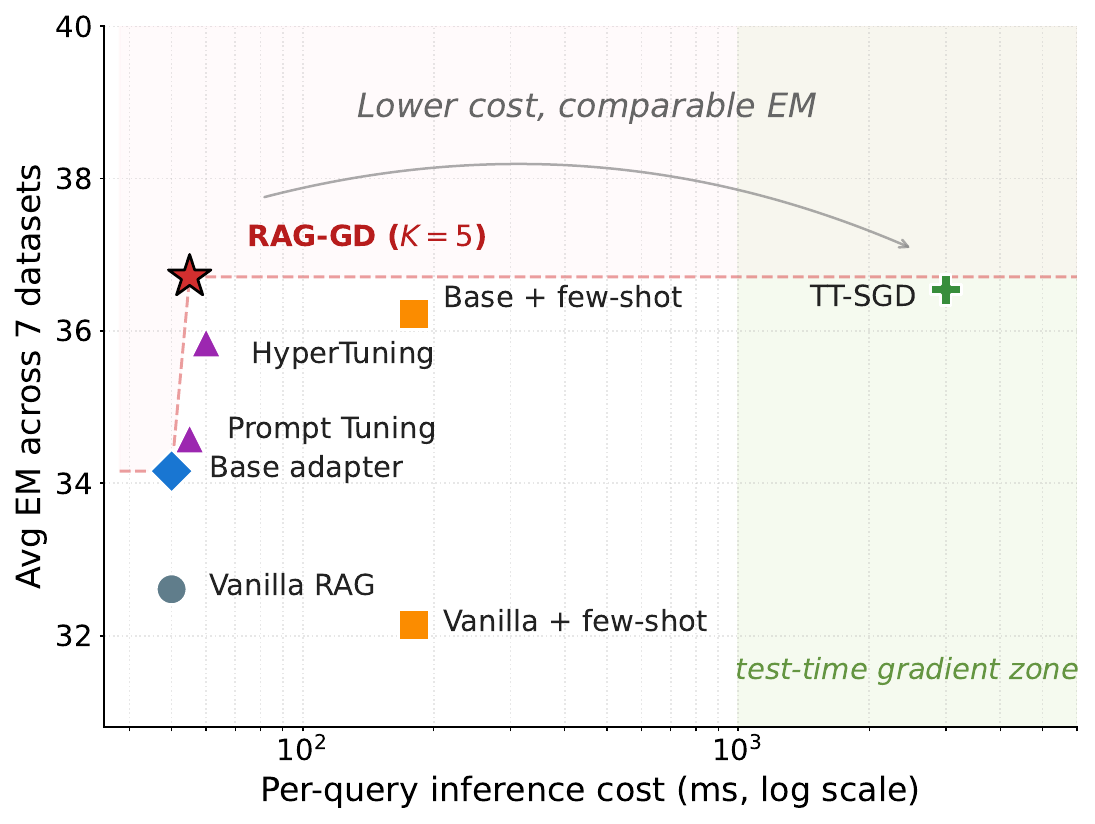}
  \caption{
  \textbf{EM-cost tradeoff} on Qwen 2.5-7B with E5 retrieval. 
  Per-query cost is shown on a log scale. 
  The shaded region marks methods that run inner GD at test time.
  }
  \label{fig:efficiency}
  \vspace{-2em}
\end{wrapfigure}

Table~\ref{tab:main} reports the headline comparison between \textsc{RAG-GD} ($K{=}5$) and the no-perturbation baselines. 
Figures~\ref{fig:family_ktrend} and~\ref{fig:efficiency} compare against additional context-conditioned methods, including Prompt Tuning, HyperTuning, TT-SGD, and + few-shot variants.
Full per-method and per-benchmark results are in Appendix~\ref{appendix:full_qa_table}, and Algorithm~\ref{alg:forward_only_2e} gives the deployment procedure.
\paragraph{The predicted update improves the base retrieval adapter.}
Across all backbone and retriever configurations, \textsc{RAG-GD} improves average EM and F1 over Base adapter. 
This comparison is controlled: both methods share the external retriever, retrieval cache, frozen backbone, and base adapter $W_0^{\mathrm{ret}}$. 
The only difference is the predicted perturbation $\widetilde{\Delta W}(C)$, which isolates the effect of adapting the generator-side retrieval interface from the support context.
\paragraph{The learned update transfers to held-out tasks.}
PopQA and Bamboogle are held out from training for both $W_0^{\mathrm{ret}}$ and $g_\phi$. 
On PopQA, \textsc{RAG-GD} improves EM over Base adapter in every backbone and retriever setting, with F1 close or improved in most cases. 
On Bamboogle, gains are strongest on Qwen, while Llama with BM25 stays close to Base adapter. 
The transfer is consistent but not uniform, suggesting that $g_\phi$ learns a reusable update rule for generator-side evidence use rather than only fitting the meta-training tasks.

\paragraph{Gradient-update supervision matters.}
Figure~\ref{fig:method_family} compares \textsc{RAG-GD} with context-conditioned baselines on Qwen 2.5-7B with E5 retrieval.
HyperTuning uses the same predictor architecture but trains through downstream task loss rather than matching the autograd-defined update. 
\textsc{RAG-GD} improves average EM and F1 over HyperTuning and Prompt Tuning, indicating that the gradient-derived target contributes beyond the predictor architecture.
\paragraph{Performance is largely insensitive to $K$.}
Figure~\ref{fig:k_trend} sweeps the inner-loop depth $K\in\{1,5,10\}$ across both backbones and retrievers. 
A single amortized step already recovers most of the gain, while additional steps bring only small and configuration-dependent changes. 
Thus, the $K{=}5$ setting in Table~\ref{tab:main} is representative. 
Full per-$K$ results are in Appendix~\ref{appendix:full_qa_table}.

\paragraph{Robustness to retrieval distractors.}
We evaluate Qwen 2.5-7B-Instruct with E5 retrieval by replacing $r\in\{0,1,3,5\}$ of the top-five documents for each query with the top-ranked document retrieved for another query from the same benchmark.
The document count and context budget are held fixed.
Replacement positions are sampled uniformly with a per-query seed, giving all methods identical corrupted contexts; support demonstrations remain clean.

\begin{table}[!htb]
\centering
\caption{Retrieval-distractor robustness over seven benchmarks. We report average EM; parentheses show retention relative to clean performance. Bold marks the highest EM at each $r$.}
\label{tab:retrieval_corruption}
\small
\setlength{\tabcolsep}{4pt}
\renewcommand{\arraystretch}{1.05}
\resizebox{0.85\textwidth}{!}{%
\begin{tabularx}{\textwidth}{l*{4}{>{\centering\arraybackslash}X}}
\toprule
\textbf{Method} & $r=0$ (clean) & $r=1$ & $r=3$ & $r=5$ \\
\midrule
Vanilla RAG & 32.61 & 30.57 (93.7\%) & 26.40 (81.0\%) & 19.12 (58.6\%) \\
Base adapter & 34.16 & 32.01 (93.7\%) & 27.52 (80.6\%) & 21.07 (61.7\%) \\
\rowcolor{ragtableblue!12}
\textcolor{ragtabletext}{\textbf{\textsc{RAG-GD} ($K{=}5$)}} & \textbf{36.71} & \textbf{34.09} (92.9\%) & \textbf{30.38} (82.8\%) & \textbf{23.88} (65.1\%) \\
\bottomrule
\end{tabularx}
}
\end{table}

\textsc{RAG-GD} achieves the highest EM at every tested level (Table~\ref{tab:retrieval_corruption}).
Its retention is slightly lower at $r=1$ but higher at $r\in\{3,5\}$.
At $r=5$, it exceeds Base adapter by 2.81 EM points, retaining 65.1\% of its clean performance versus 61.7\% for Base adapter.
Robustness to passages explicitly supporting incorrect answers requires separate evaluation.

\paragraph{Amortization approaches test-time adaptation at lower cost.}
Figure~\ref{fig:efficiency} shows the EM-cost tradeoff for Qwen 2.5-7B with E5 retrieval.
TT-SGD performs inner-loop backpropagation through the 7B LLM at test time, while \textsc{RAG-GD} moves this computation into training and uses only one forward pass through $g_\phi$ at inference. 
As a result, \textsc{RAG-GD} reaches a similar average EM and F1 operating point at substantially lower per-query cost.

\subsection{Discussion}
\label{sec:exp-discussion}

These results complete the theory-to-practice arc. 
The linear regime gives an exact gradient-descent correspondence, while the nonlinear experiments show where this correspondence becomes feature-dependent. 
At LLM scale, we do not claim that frozen RAG LLMs implement the linear equivalence internally. 
Instead, the autograd-defined update provides a practical target for context-conditioned adaptation. 
Because \textsc{RAG-GD} and Base adapter share the same retriever, retrieval cache, backbone, and $W_0^{\mathrm{ret}}$, the gains isolate the contribution of the predicted update to the generator-side retrieval interface. 
The held-out transfer and cost-performance tradeoff suggest that gradient-supervised retrieval-interface adaptation is a promising forward-only alternative to test-time backpropagation for RAG.

\section{Conclusion}

We studied retrieval-augmented generation through an in-context optimization lens. 
In a controlled linear setting, we showed that one linear self-attention layer can implement one gradient-descent step on a unified linearized RAG loss covering projection-based and dot-product retrieval interfaces. 
Empirical tests verified this construction in the exact regime and revealed a structured boundary under nonlinear architectures and real regression data, where alignment becomes sensitive to feature distribution. 
At LLM scale, we turned this view into a forward-only adaptation method by using the autograd-defined $K$-step update to a generator-side Q/K/V LoRA interface as supervision for a lightweight predictor. 
Across seven QA benchmarks, two retrievers, and two frozen backbones, the predicted update improved a shared-adapter baseline, transferred to held-out domains, and was largely insensitive to $K$. 
Overall, these results suggest that retrieved evidence can be treated not only as external context, but also as a signal for context-induced adaptation in RAG.
\section{Limitations}
\label{sec:limitations} 
Our linear construction is an analytical starting point rather than a literal account of modern RAG LLMs, and the nonlinear experiments show that the correspondence depends on architecture and feature distribution. 
At LLM scale, we instantiate the view with a generator-side Q/K/V LoRA interface while keeping the retriever and backbone fixed. 
Open questions remain about the inner-loop optimizer, update parameterization, predictor capacity, scaling to larger backbones, and robustness across broader retrieval settings. 
Future work should study whether larger support sets can be aggregated into parameter updates to reduce the prompt-length overhead of concatenated ICL, while accounting for support-encoding cost and information loss in mean pooling.
Further directions include combining predicted updates with selectively retained demonstrations and using uncertainty-aware gating to suppress unreliable updates under noisy retrieval.

\section{Acknowledgments}
This work was supported in part by the National Institutes of Health (NIH; grants 5R01DA056470 and 5R01AG080670) and the U.S. Department of Veterans Affairs Health Systems Research (HSR; awards 5I01HX003711 and 1I01HX003969). The content is solely the responsibility of the authors and does not necessarily represent the official views of the NIH or the U.S. Department of Veterans Affairs.

\bibliography{references}
\bibliographystyle{plain}

\section*{Broader Impacts}
The method we propose lowers the cost of adapting a deployed large language model to a new task or domain. At inference, the LLM still runs a forward pass to generate the answer, but no backward pass through the LLM is required: adapting to the context costs only a single small forward pass through a context-conditional weight predictor. If adopted at scale, this could reduce the energy footprint of test-time adaptation pipelines for retrieval-augmented systems. The flip side is that lower adaptation cost could also accelerate the deployment of models in domains for which the underlying LLM has not been carefully evaluated, including settings where retrieval can amplify biases in the corpus. We encourage practitioners adopting this style of adaptation to retain the same evaluation discipline that would apply to a fine-tuned model. All datasets used in our evaluation are publicly available QA and regression benchmarks that do not contain personally identifiable or sensitive information, and our work does not raise additional ethical concerns beyond those discussed above.

\section*{Reproducibility Statement}
All datasets used in our main evaluation are publicly available QA benchmarks (NQ, TriviaQA, PopQA, HotpotQA, 2WikiMultiHopQA, MuSiQue, Bamboogle). The supplementary regression datasets discussed in the appendix are also public. Preprocessing steps, dataset splits, and the training pools used for the reference retrieval adapter $W_0^{\text{ret}}$ and the predictor $g_\phi$ are documented in the appendix.

Our implementation uses PyTorch with the Hugging Face Transformers library on top of frozen Qwen 2.5-7B-Instruct~\citep{yang2024qwen} and Llama 3.1-8B-Instruct~\citep{dubey2024llama} backbones. Hyperparameters for the reference retrieval adapter $W_0^{\text{ret}}$, the predictor $g_\phi$, and the inner-loop SGD target ($\eta$, $K$, number of demonstrations $N$) are listed in Appendix~\ref{app:impl}. All experiments were conducted on NVIDIA A100 GPUs.

%%%%%%%%%%%%%%%%%%%%%%%%%%%%%%%%%%%%%%%%%%%%%%%%%%%%%%%%%%%%

\appendix
\setcounter{equation}{0}
\renewcommand{\theequation}{\arabic{equation}}
% Keep appendix equation anchors distinct from the main text after the reset.
\renewcommand{\theHequation}{appendix.\arabic{equation}}

\section{Proof of Lemma~\ref{lemma:equivalence}}\label{Appendix: Proposition1}

\paragraph{Statement.}
Given a 1-head linear-attention layer and tokens $e_j = (x_1^j,\, x_2^j,\, y^j)$ for $j = 1,\ldots,N$, we construct key, query, and value matrices $W_K$, $W_Q$, $W_V$ and a projection $P$ such that one linear self-attention update on each $e_j$ matches one gradient-descent step on the unified RAG loss of Section~\ref{sec:method}. The update modifies only the $y$-coordinate of each token:
\begin{equation}
e_j \leftarrow e_j + \big(0,\; 0,\; \Delta W_1 x_1^j + \Delta W_2 x_2^j\big) \;=\; e_j + P\, V\, K^\top q_j,
\label{eq:appA_update}
\end{equation}
where $\Delta W_1$ and $\Delta W_2$ are the gradient-step updates of Eq.~\ref{eq:gd_step} in the main text.

\paragraph{Setup.}
We are given $N$ context tokens, each of the form $e^i = (x_1^i,\, x_2^i,\, y^i)$ corresponding to one training pair, plus a query token $e_{N+1} = (x_1^{N+1},\, x_2^{N+1},\, 0)$ at position $N{+}1$. The model is asked to predict the updated $y$-value at position $N{+}1$.

\paragraph{Step 1: expand the post-update prediction.}
Writing the post-update prediction $y'$ as the original prediction plus the contributions of $\Delta W_1$ and $\Delta W_2$,
\begin{align}
y' &= W_1' x_1 + W_2' x_2 \nonumber \\
   &= (W_1 + \Delta W_1)\,x_1 + (W_2 + \Delta W_2)\,x_2 \nonumber \\
   &= W_1 x_1 + W_2 x_2 \;+\; \Delta W_1 x_1 + \Delta W_2 x_2.
\label{eq:appA_yprime}
\end{align}

\paragraph{Step 2: gradient step on the unified RAG loss.}
Under the squared loss $L(W_1, W_2) = \tfrac{1}{2N}\sum_{i=1}^N \|W_1 x_1^i + W_2 x_2^i - y_i\|^2$, one gradient step with learning rate $\eta$ yields
\begin{align}
\Delta W_1 &= -\eta\,\nabla_{W_1} L = -\frac{\eta}{N}\sum_{i=1}^N \big(W_1 x_1^i + W_2 x_2^i - y_i\big)\,(x_1^i)^\top, \label{eq:appA_dW1}\\
\Delta W_2 &= -\eta\,\nabla_{W_2} L = -\frac{\eta}{N}\sum_{i=1}^N \big(W_1 x_1^i + W_2 x_2^i - y_i\big)\,(x_2^i)^\top, \label{eq:appA_dW2}\\
\Delta y &= \Delta W_1 x_1 + \Delta W_2 x_2.
\label{eq:appA_dy}
\end{align}

\paragraph{Step 3: rewrite $\Delta y$ as a sum of outer-product contractions.}
Substituting Eqs.~\ref{eq:appA_dW1}--\ref{eq:appA_dW2} into Eq.~\ref{eq:appA_dy} and evaluating at the query token $j$,
\begin{align}
\Delta y \;=\; -\frac{\eta}{N}\sum_{i=1}^N \big(W_1 x_1^i + W_2 x_2^i - y_i\big)\,(x_1^i)^\top x_1^j \;-\; \frac{\eta}{N}\sum_{i=1}^N \big(W_1 x_1^i + W_2 x_2^i - y_i\big)\,(x_2^i)^\top x_2^j.
\label{eq:appA_dy_expanded}
\end{align}
Equivalently, the update applied to the token at position $j$ is
\begin{align}
\begin{pmatrix} x_1^j \\ x_2^j \\ y^j \end{pmatrix} \;\leftarrow\; \begin{pmatrix} x_1^j \\ x_2^j \\ y^j \end{pmatrix} \;+\; \begin{pmatrix} 0 \\ 0 \\ \Delta y \end{pmatrix},
\qquad \text{with}\qquad \begin{pmatrix} 0 \\ 0 \\ \Delta y \end{pmatrix} = \begin{pmatrix} 0 \\ 0 \\ \Delta W_1 x_1 + \Delta W_2 x_2 \end{pmatrix}.
\label{eq:appA_token_update}
\end{align}

\paragraph{Step 4: cast the update as a linear self-attention output.}
Using the identity $a\, b^\top c = (a \otimes b^\top)\, c$ and grouping terms, Eq.~\ref{eq:appA_dy_expanded} can be written as
\begin{align}
\begin{pmatrix} 0 \\ 0 \\ \Delta y \end{pmatrix}
\;=\; -\frac{\eta}{N}\sum_{i=1}^N
\underbrace{\begin{pmatrix} 0 \\ 0 \\ W_1 x_1^i + W_2 x_2^i - y^i \end{pmatrix}}_{\text{value vector } v_i}
\,\otimes\,
\underbrace{\begin{pmatrix} x_1^i & x_2^i & 0 \end{pmatrix}}_{\text{key vector } k_i^\top}
\;\underbrace{\begin{pmatrix} x_1^j \\ x_2^j \\ 0 \end{pmatrix}}_{\text{query vector } q_j}.
\label{eq:appA_outer}
\end{align}
Each factor in Eq.~\ref{eq:appA_outer} can be obtained by applying a fixed linear projection to the token $e^i = (x_1^i, x_2^i, y^i)$ or $e^j$:
\begin{align}
v_i &= \underbrace{\begin{pmatrix}
0 & 0 & 0\\
0 & 0 & 0\\
W_1 & W_2 & -I_y
\end{pmatrix}}_{W_V}\, e^i, \qquad
k_i = \underbrace{\begin{pmatrix}
I_x & 0 & 0\\
0 & I_x & 0\\
0 & 0 & 0
\end{pmatrix}}_{W_K}\, e^i, \qquad
q_j = \underbrace{\begin{pmatrix}
I_x & 0 & 0\\
0 & I_x & 0\\
0 & 0 & 0
\end{pmatrix}}_{W_Q}\, e^j.
\label{eq:appA_proj}
\end{align}

\paragraph{Step 5: explicit construction.}
Combining Eqs.~\ref{eq:appA_outer} and~\ref{eq:appA_proj}, the gradient step of Eqs.~\ref{eq:appA_dW1}--\ref{eq:appA_dW2} is realized as a linear self-attention update with the closed-form projections
\begin{equation}
\resizebox{0.9\linewidth}{!}{$
\begin{pmatrix}
x_1^j\\
x_2^j\\
y^j
\end{pmatrix}
\leftarrow
\begin{pmatrix}
x_1^j\\
x_2^j\\
y^j
\end{pmatrix}
-\frac{\eta}{N}\sum_{i=1}^{N}
\Bigg(
\underbrace{\begin{pmatrix}
0&0&0\\
0&0&0\\
W_1&W_2&-I_y
\end{pmatrix}}_{W_V}
\begin{pmatrix}
x_1^i\\
x_2^i\\
y^i
\end{pmatrix}
\Bigg)
\otimes
\Bigg(
\underbrace{\begin{pmatrix}
I_x&0&0\\
0&I_x&0\\
0&0&0
\end{pmatrix}}_{W_K}
\begin{pmatrix}
x_1^i\\
x_2^i\\
y^i
\end{pmatrix}
\Bigg)^\top
\Bigg(
\underbrace{\begin{pmatrix}
I_x&0&0\\
0&I_x&0\\
0&0&0
\end{pmatrix}}_{W_Q}
\begin{pmatrix}
x_1^j\\
x_2^j\\
y^j
\end{pmatrix}
\Bigg)
$}
\label{con:main_pro}
\end{equation}
The projection $P$ is taken to be the identity on the $y$-coordinate, so that the value contribution lands in the $y$-slot of $e^j$. Comparing the right-hand side of Eq.~\ref{con:main_pro} to the gradient step in Eqs.~\ref{eq:appA_dW1}--\ref{eq:appA_dW2}, the two sides agree term by term. One linear self-attention update therefore reproduces one gradient-descent step on the unified RAG loss, as claimed. \qed

\section{Linear RAG: Derivation of the Unified Retriever Formulation}\label{appendix:linearRAG}

\paragraph{Main function.}
\begin{equation}
y = (W_q, W_z)
\begin{bmatrix}
x_q \\[3pt]
\sum_{i=1}^n (W_e x_q)^\top (W_e d_i)\, d_i
\end{bmatrix}
= W_q x_q + W_z \sum_{i=1}^n (W_e x_q)^\top (W_e d_i)\, d_i.
\end{equation}

\paragraph{Define $M$ and rewrite the similarity.}
We adopt the shared-encoder simplification, where the same linear encoder $W_e$ maps both queries and documents into the retrieval space. The general DPR formulation~\citep{karpukhin2020dense} permits separate $W_q$ and $W_d$, in which case the analysis below carries through with $M = W_q^\top W_d$ but $M$ need not be symmetric. Defining
\begin{equation}
M \triangleq W_e^\top W_e
\;\;\;\;\Rightarrow\;\;\;\;
(W_e x_q)^\top (W_e d_i) = x_q^\top W_e^\top W_e d_i = x_q^\top M d_i.
\end{equation}
Hence,
\begin{equation}
y = W_q x_q + W_z \sum_{i=1}^n (x_q^\top M d_i)\, d_i.
\end{equation}

\paragraph{Converting ``scalar $\times$ vector'' into ``matrix $\times$ vector.''}
Note that $x_q^\top M d_i$ is a scalar, and the following identity holds:
\begin{equation}
(x_q^\top M d_i)\, d_i = d_i (d_i^\top M^\top x_q) = (d_i d_i^\top) M^\top x_q.
\end{equation}
Therefore,
\begin{equation}
\sum_{i=1}^n (x_q^\top M d_i)\, d_i
= \sum_{i=1}^n (d_i d_i^\top) M^\top x_q
= \Big(\sum_{i=1}^n d_i d_i^\top\Big) M^\top x_q.
\end{equation}

\paragraph{Define the document second-moment matrix $D$.}
\begin{equation}
D \triangleq \sum_{i=1}^n d_i d_i^\top
\;\;\;\;\Rightarrow\;\;\;\;
\sum_{i=1}^n (x_q^\top M d_i)\, d_i
= D M^\top x_q.
\end{equation}

\paragraph{Substituting back into $y$.}
\begin{equation}
y = W_q x_q + W_z D M^\top x_q.
\end{equation}

Then $M = W_e^\top W_e$ is symmetric, i.e., $M^\top = M$. Thus the expression simplifies to
\begin{equation}
y = W_q x_q + W_z D M x_q.
\end{equation}

The right-hand side is grouped into an equivalent linear mapping:
\begin{equation}
y = (W_q + W_z D M)\, x_q.
\end{equation}

\section{Linear-Regime Equivalence: Full Setup and Additional Figures}\label{appendix:linear}

This appendix supplements Section~\ref{sec:linear_emp} of the main paper with a full description of the synthetic-regression setup and the additional robustness and stacked-layer figures referenced there.

\subsection{Setup Details}

\paragraph{Tokens.}
Each token concatenates an input vector, a retrieval-derived feature, and a target,
\begin{equation}
e_i = (x_i,\, z_i,\, y_i), \qquad i = 1, \ldots, N,
\end{equation}
where $N$ is the number of in-context examples for a single task $\tau$. The auxiliary slot $z_i$ instantiates the unified RAG view of Section~\ref{sec:method}. Under the linear-projection retriever, $z_i$ is a document-derived feature; under the dot-product retriever, $z_i = x_i$ and the document information is injected through the keys and values rather than through the token (see ``dot-product injection'' below).

\paragraph{Pre-training objective.}
We train an LSA layer parameterized by $\theta$ to minimize the expected squared prediction error across tasks:
\begin{equation}
\mathcal{L}(\theta) = \frac{1}{B} \sum_{\tau=1}^{B}
\big\| \hat{y}_{\theta}\big( \{e_{\tau, i}\}_{i=1}^{N},\; e_{\tau, N+1} \big) - y_{\tau, N+1} \big\|^2,
\end{equation}
where the query token at position $N+1$ is $e_{\tau, N+1} = (x_{\text{test}}, z_{\text{test}}, 0)$ and $y_{\tau, N+1}$ is its target. The objective is optimized with minibatch SGD over a fresh batch of tasks at each iteration. We denote the parameters at convergence by $\theta^{*}$.

\paragraph{Synthetic data.}
Following~\citep{garg2022can, von2023transformers}, we generate each task $\tau$ from a teacher with weights $W_\tau \sim \mathcal{N}(0, I)$. Inputs are drawn from $x_{\tau, i} \sim \mathcal{U}(-1, 1)^{n_I}$ and targets are constructed as $y_{\tau, i} = W_\tau^{1}\, x_{\tau, i}^{1} + W_\tau^{2}\, x_{\tau, i}^{2}$. We set $N = n_I = 10$ with output dimension $1$, and we sweep the document count $k \in \{2, 5, 10, 25\}$.

\paragraph{Dot-product injection.}
Under the linear-projection retriever the document is included in the token, $e_i = (x_i, \mathcal{D}, y_i)$, and the LSA layer can learn to select relevant documents during pre-training. Under the dot-product retriever the document is not concatenated into the token; instead, document information is injected directly into the key and value matrices,
\begin{equation}
K = \begin{bmatrix} K_{\text{ctx}} \\ h_d \end{bmatrix},
\qquad
V = \begin{bmatrix} V_{\text{ctx}} \\ h_d \end{bmatrix},
\end{equation}
where $K_{\text{ctx}}, V_{\text{ctx}}$ are the contextual key/value rows and $h_d = f(\mathcal{D}) \in \mathbb{R}^{B \times \dim(x)}$ is a fixed projection of the document set $\mathcal{D} = \{d_1, \ldots, d_n\}$ into the input dimension.

\paragraph{Constructed reference predictor.}
The trained LSA layer (parameters $\theta^{*}$) is compared against a \emph{constructed} predictor that realizes one gradient-descent step on the unified RAG loss exactly. Following the construction of Eq.~\ref{con:main_pro}, we set the value, key, and query projections so that one LSA update reproduces the gradient step. For the linear-projection retriever, $W_1$ and $W_2$ inside $W_V$ are initialized to zero, following~\citep{von2023transformers}. For the dot-product retriever, $W_2 = W_z\big(\sum_i d_i d_i^\top\big) M^\top$, with $W_z \in \mathbb{R}^{d_y \times d_d}$ and $M \in \mathbb{R}^{d_d \times d_d}$ sampled independently from $\mathcal{N}(0, \sigma^2)$, and document features $C \sim \mathcal{U}(-\tfrac{1}{2}, \tfrac{1}{2})^{k \times d_d}$. The inner-loop learning rate $\eta$ for the constructed predictor is chosen by line search to minimize the constructed model's loss over $10^4$ training tasks. We write the resulting prediction $\hat{y}_{\theta, \mathrm{rag}}(x_{\text{test}})$.

\paragraph{Evaluation metrics.}
On $T_{\text{val}} = 10^4$ held-out validation tasks, following~\citep{von2023transformers}, we report the mean of three quantities between the trained and constructed predictors: (i) the prediction difference $\big\| \hat{y}_{\theta^{*}}(x_{\tau, \text{test}}) - \hat{y}_{\theta, \mathrm{rag}}(x_{\tau, \text{test}}) \big\|_2$; (ii) the cosine similarity between the input-sensitivities $\partial \hat{y}_{\theta, \mathrm{rag}}/\partial x_{\text{test}}$ and $\partial \hat{y}_{\theta^{*}}/\partial x_{\text{test}}$; and (iii) the corresponding $\ell_2$ sensitivity difference.

\subsection{Robustness to Distribution Shift and Document Count}

\begin{figure*}[t]
    \centering
    \includegraphics[width=0.24\textwidth]{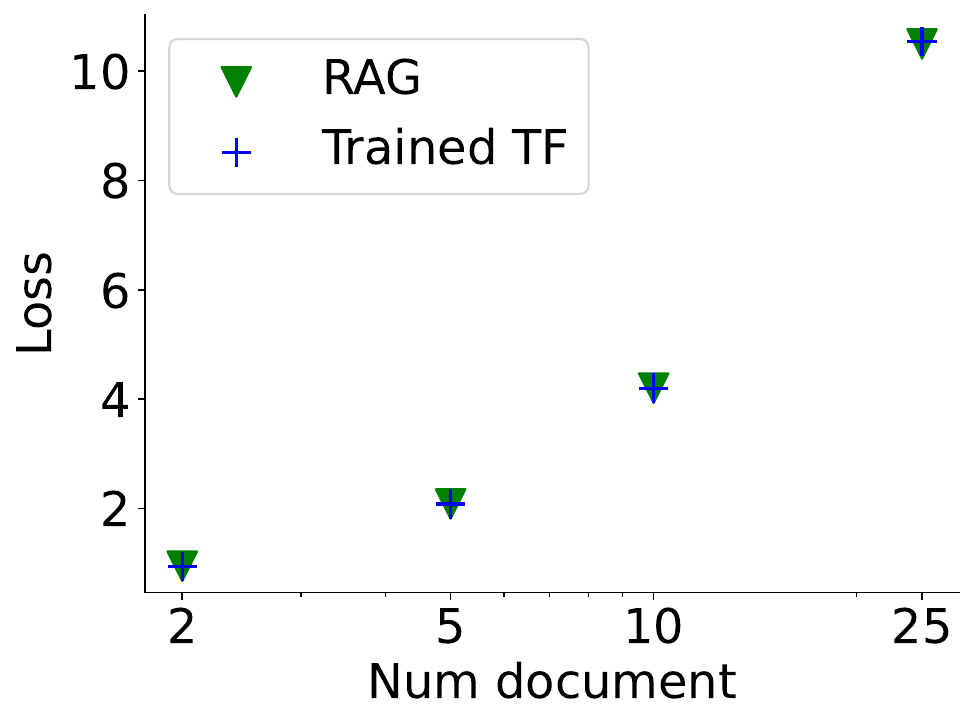}%
    \includegraphics[width=0.24\textwidth]{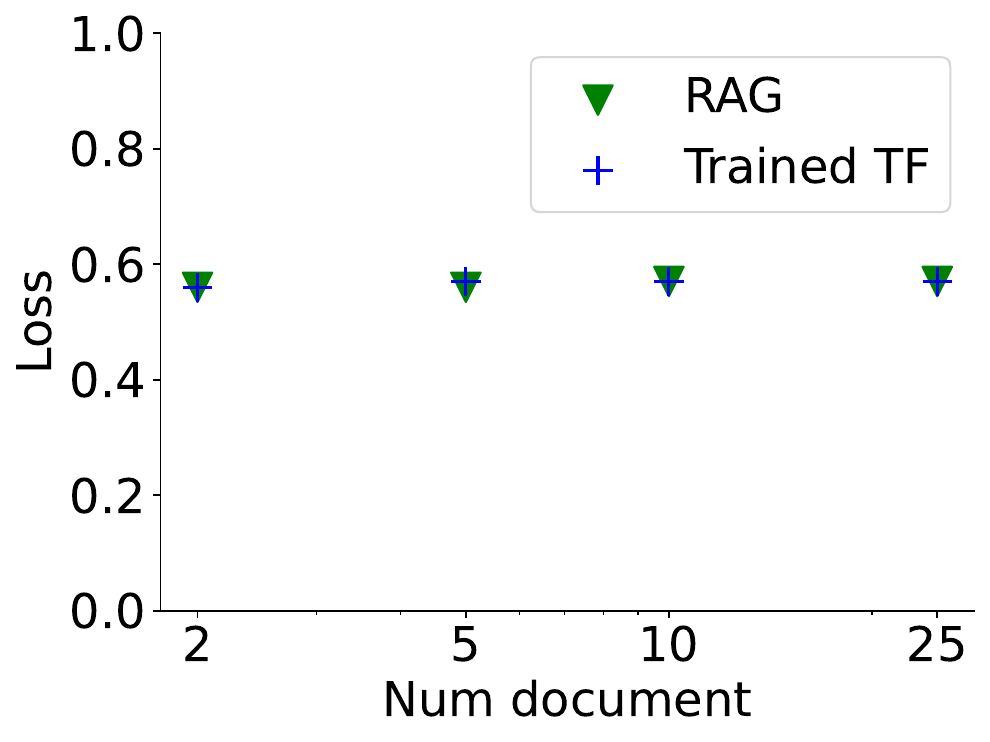}%
    \includegraphics[width=0.24\textwidth]{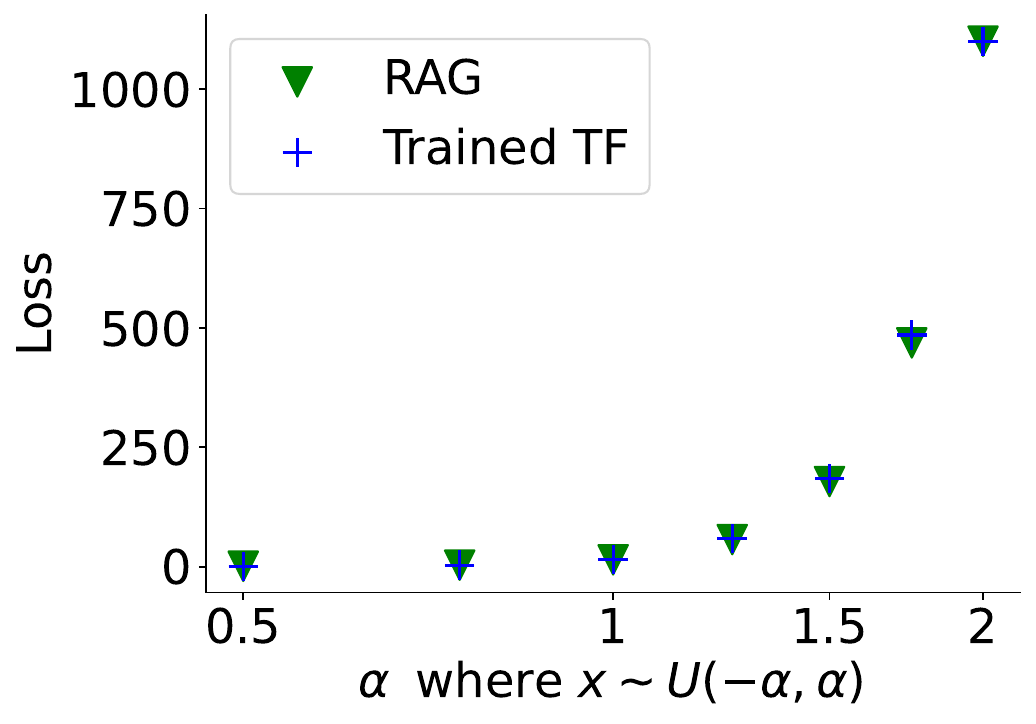}%
    \includegraphics[width=0.24\textwidth]{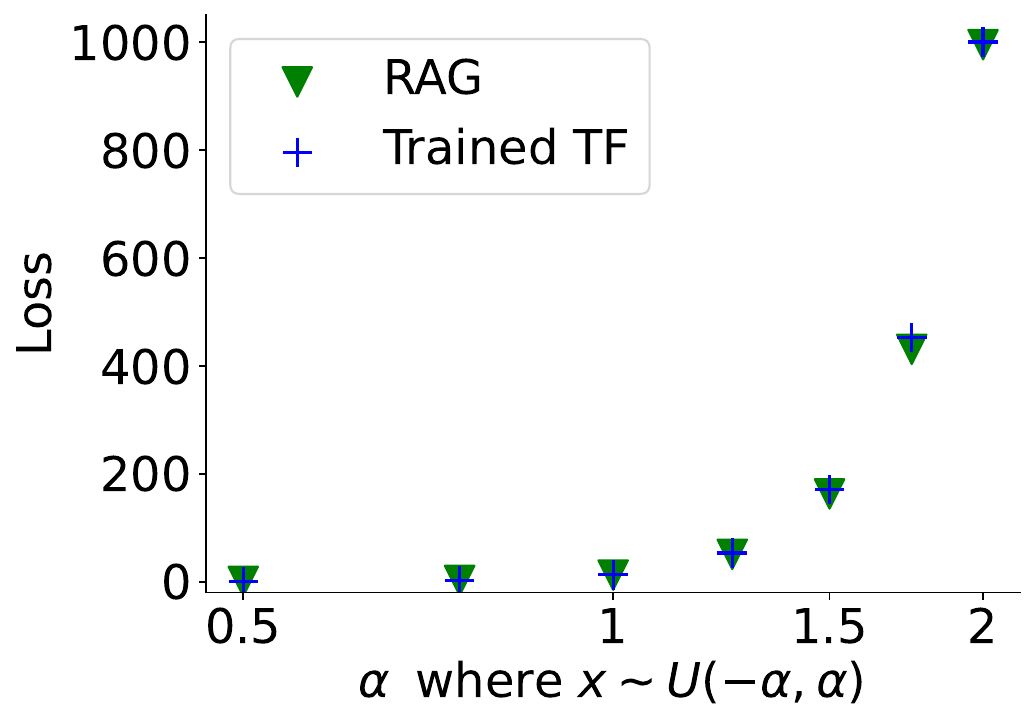}%
    \caption{Robustness of the single-layer agreement of Section~\ref{sec:linear_emp}. \textit{Left two:} loss as a function of document count for the linear-projection (left) and dot-product (center-left) retrievers. \textit{Right two:} loss under input distribution shift, with test inputs drawn from $\mathcal{U}(-\alpha, \alpha)^{n_I}$ for varying $\alpha$, for the linear-projection (center-right) and dot-product (right) retrievers. The trained Transformer, the constructed gradient-descent predictor, and their interpolation track each other closely in all settings.}
    \label{figure:differ_doc_distrubution}
\end{figure*}

To probe whether the trained LSA layer captures a generalizable update rule rather than memorizing the training distribution, we vary two factors at test time. First, we sweep the document count $n \in \{2, 5, 10, 25\}$ and recompute the comparison; second, we sample test inputs from $\mathcal{U}(-\alpha, \alpha)^{n_I}$ with $\alpha \in \{0.5, 1, 1.5, 2\}$ while training is fixed at $\alpha = 1$. Figure~\ref{figure:differ_doc_distrubution} reports the resulting loss curves. With the linear-projection retriever, the absolute loss rises with document count (the projected document features carry more variance) but the LSA layer follows the gradient predictor in lockstep. With the dot-product retriever, where the document information enters through the second-moment matrix $\sum_i d_i d_i^\top$, the loss is largely insensitive to document count. The dot-product variant is also computationally cheaper, since no per-document projection is required.

\subsection{Stacked-Layer Agreement under the Dot-Product Retriever}

\begin{figure*}[t]
    \centering
    \includegraphics[width=0.24\textwidth]{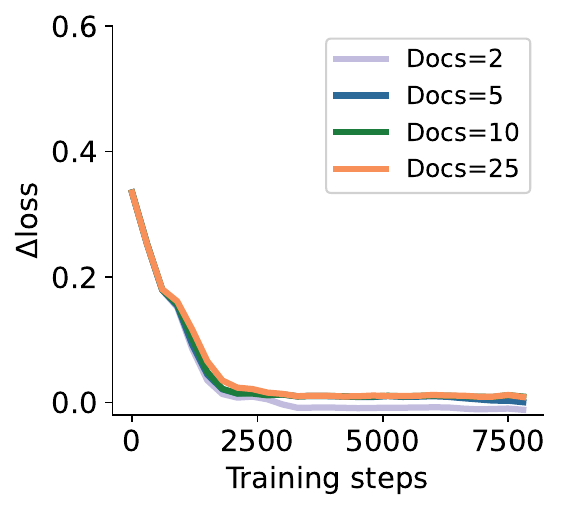}%
    \includegraphics[width=0.24\textwidth]{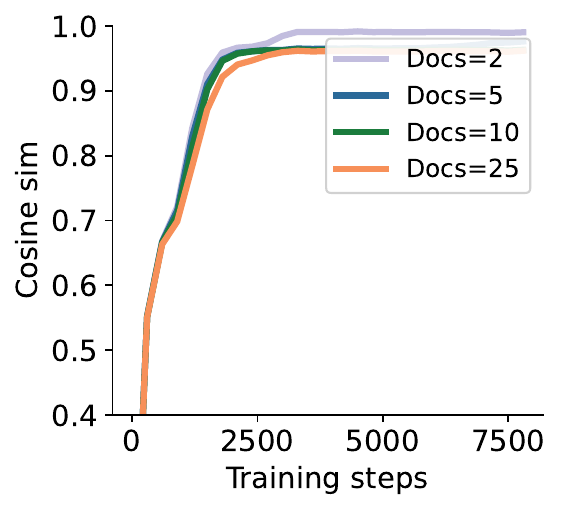}%
    \includegraphics[width=0.24\textwidth]{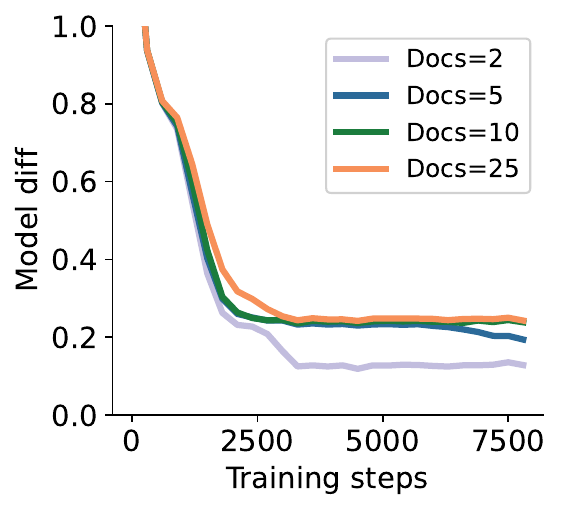}%
    \includegraphics[width=0.24\textwidth]{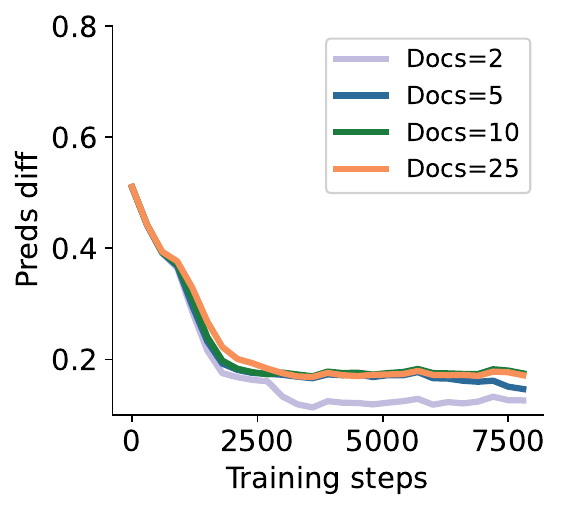}\\[1ex]
    \includegraphics[width=0.24\textwidth]{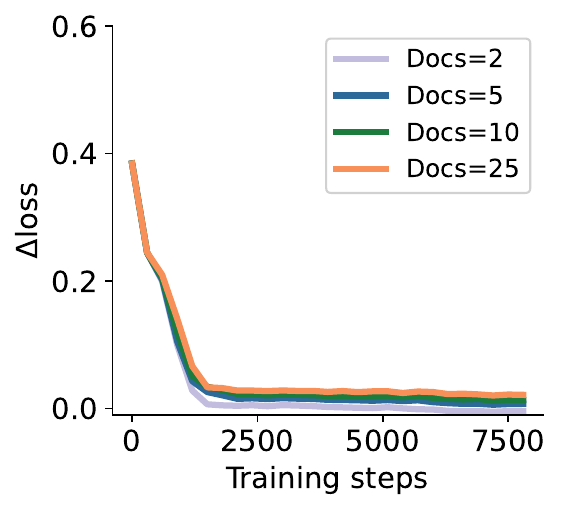}%
    \includegraphics[width=0.24\textwidth]{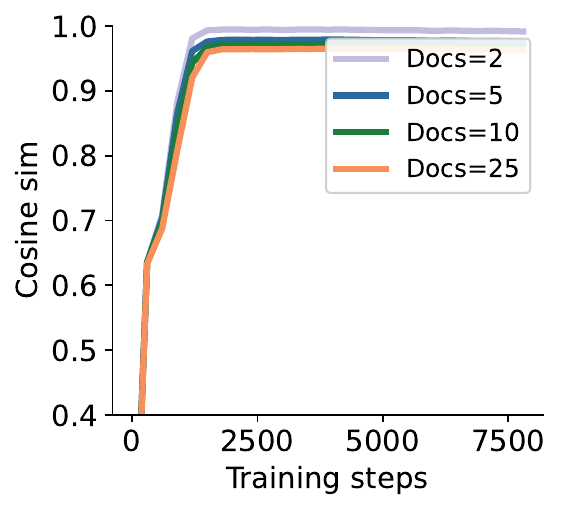}%
    \includegraphics[width=0.24\textwidth]{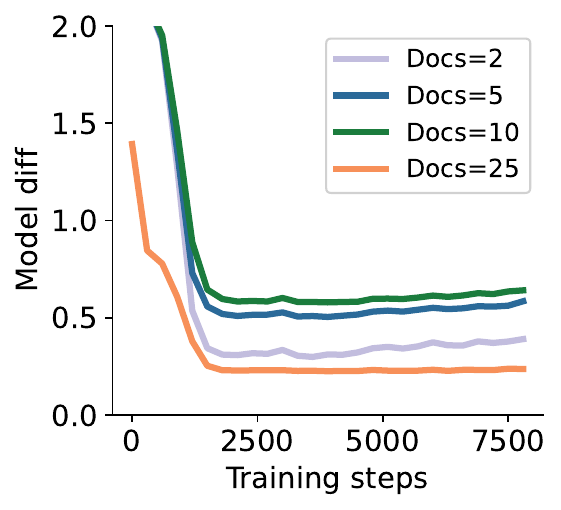}%
    \includegraphics[width=0.24\textwidth]{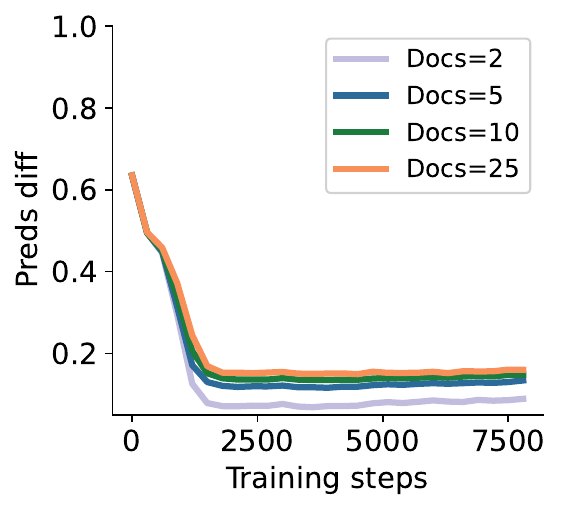}%
    \caption{Stacked-layer agreement under the dot-product retriever (Section~\ref{sec:linear_stress}). \textit{Top row:} 2-layer model. \textit{Bottom row:} 5-layer model. Columns: (a) loss difference between trained Transformer and constructed gradient-descent predictor, (b) sensitivity cosine, (c) model difference, (d) prediction difference. Agreement remains close at both depths; the small residual gap at $\text{Docs}=25$ in the 2-layer setting shrinks as depth increases to 5.}
    \label{figure:layer_different_doc_R2}
\end{figure*}

Figure~\ref{figure:layer_different_doc_R2} reports the dot-product variant at depths 2 and 5. The loss differences between the trained Transformer and the constructed predictor remain small across document counts, and the prediction differences converge to similar values. The number of retrieved documents has a depth-dependent effect on the residual: at depth 2, the prediction difference is smaller for $\text{Docs}=2$ than for $\text{Docs}=25$, but this gap narrows at depth 5. The corresponding analysis for the linear-projection retriever is reported in Appendix~\ref{section:RAG_non_linear_layers}.

\section{Normalization Analysis: Per-Dataset Extended Results}\label{appendix:normalization}

This appendix supplements Section~\ref{sec:nonlinear_emp} of the main paper. Section~\ref{sec:nonlinear_emp} reports the headline result on Bike Sharing and California Housing; here we cover the remaining two datasets, Predict Calorie Expenditure and Wine Quality. Datasets and normalization methods are as defined in Section~\ref{sec:nonlinear_emp} and Appendix~\ref{con:dataset_detals}. The training set is used as the retrieval corpus and is normalized with Z-score throughout; only the input-side normalization is varied, between Z-score~\citep{bishop2006pattern}, Min--Max~\citep{bishop2006pattern}, rank-based~\citep{conover1999practical}, and Tanh~\citep{maaten2008visualizing}.

On Predict Calorie Expenditure, the trained Transformer continues to track the gradient-descent predictor closely, mirroring the alignment seen on Bike Sharing in Section~\ref{sec:nonlinear_emp}. Wine Quality is the harder case. Under Min--Max normalization, a few outliers dominate the scaling and compress most of the samples near zero. Two effects follow. The sensitivity cosine drops because the sensitivity vectors diverge from those of the gradient-descent predictor. The prediction difference also fluctuates more strongly, indicating instability in the alignment between RAG and ICL dynamics under heavy-tailed feature distributions. This is consistent with the California Housing pattern in Section~\ref{sec:nonlinear_emp}: when retrieval-derived dot products are dominated by skewed features, the linear correspondence becomes less predictive.

\begin{figure*}[t]
    \centering
    \includegraphics[width=0.24\textwidth]{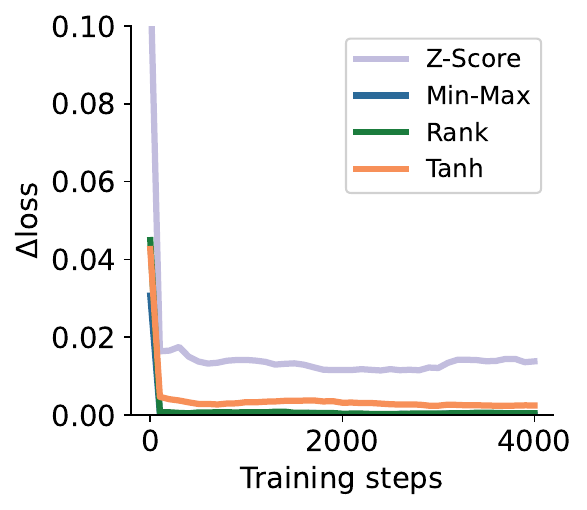}%
    \includegraphics[width=0.24\textwidth]{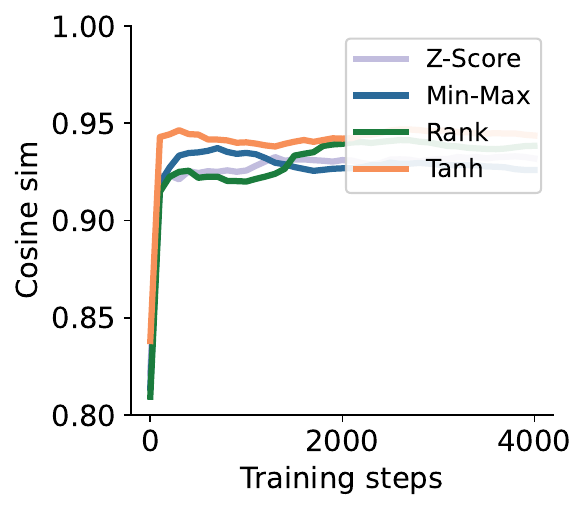}%
    \includegraphics[width=0.24\textwidth]{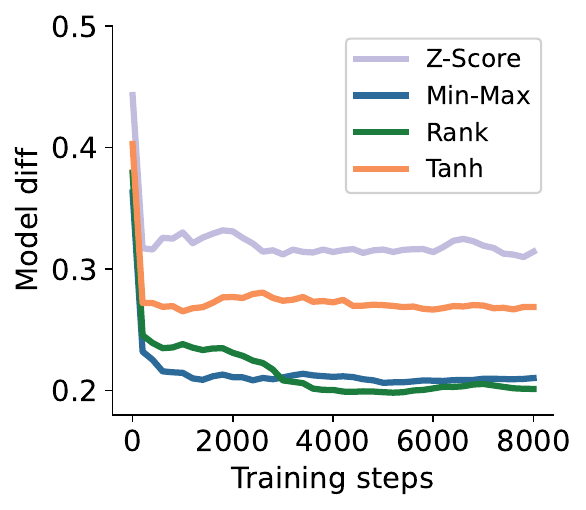}%
    \includegraphics[width=0.24\textwidth]{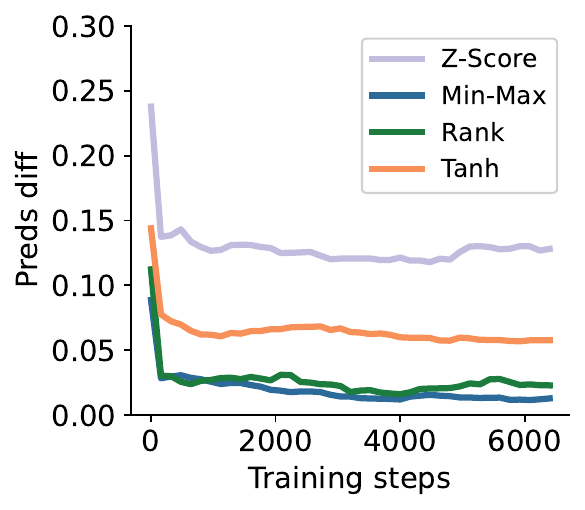}\\[1ex]
    \includegraphics[width=0.24\textwidth]{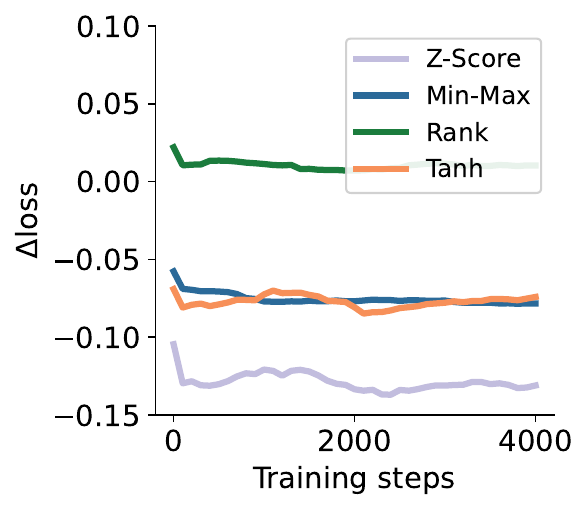}%
    \includegraphics[width=0.24\textwidth]{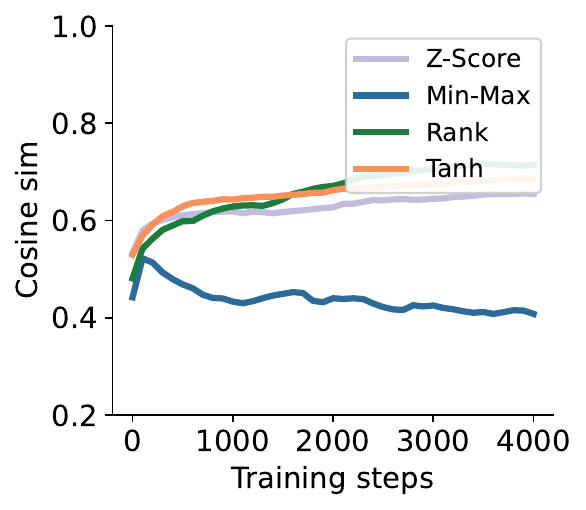}%
    \includegraphics[width=0.24\textwidth]{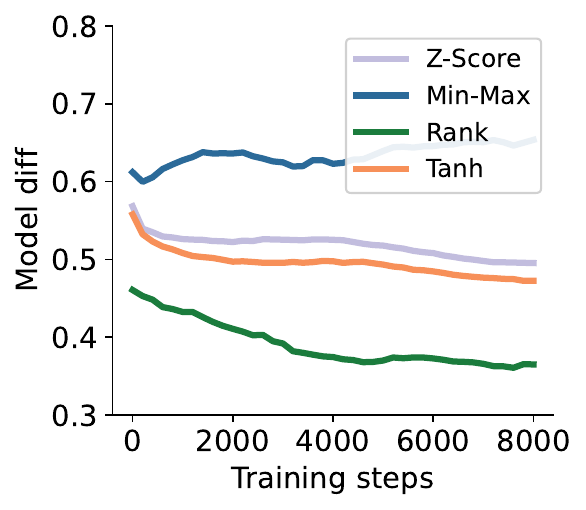}%
    \includegraphics[width=0.24\textwidth]{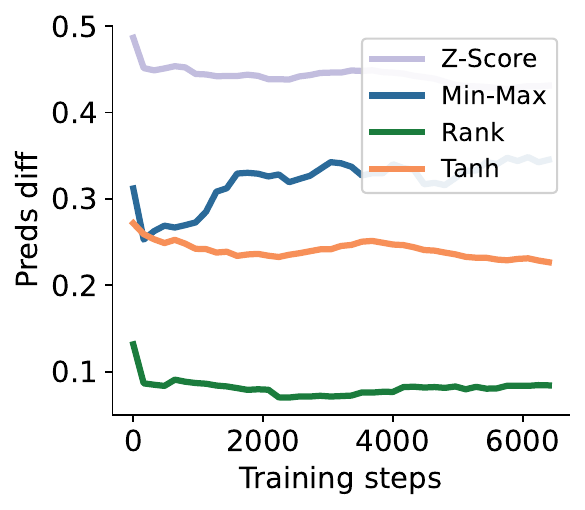}%
    \caption{Per-dataset normalization results. \textit{Top row:} Predict Calorie Expenditure. \textit{Bottom row:} Wine Quality. From left to right, the columns report loss difference with the trained Transformer, sensitivity cosine, model difference, and prediction difference. The four normalization schemes (Z-score, Min--Max, rank-based, Tanh) are overlaid within each panel.}
    \label{figure:cals_win}
\end{figure*}

\section{Stacked-Layer Agreement under the Projection-Based Retriever}

\label{section:RAG_non_linear_layers}
\begin{figure*}[t]
    \centering
    \includegraphics[width=0.24\textwidth]{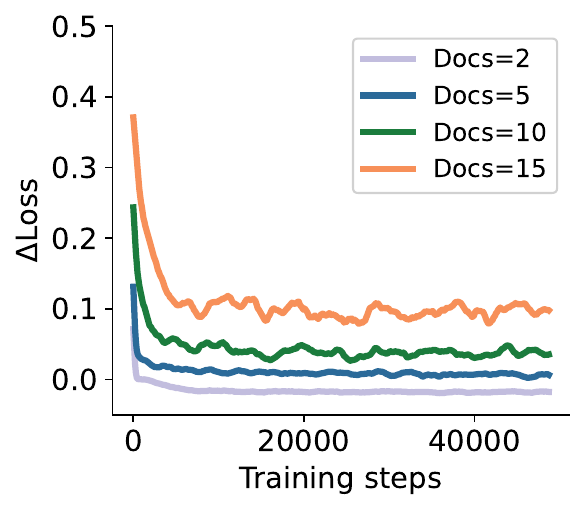}%
    \includegraphics[width=0.24\textwidth]{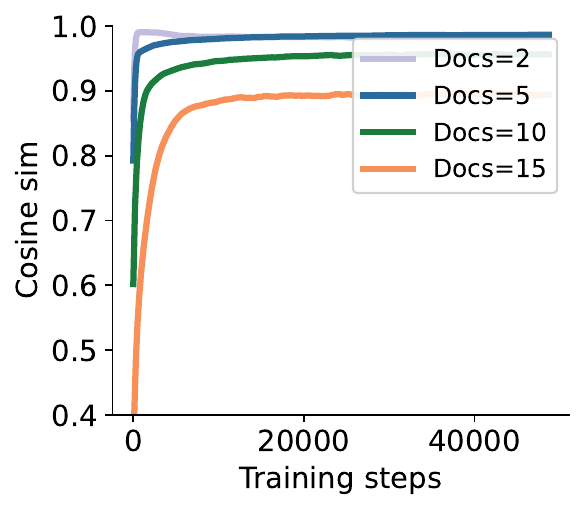}%
    \includegraphics[width=0.24\textwidth]{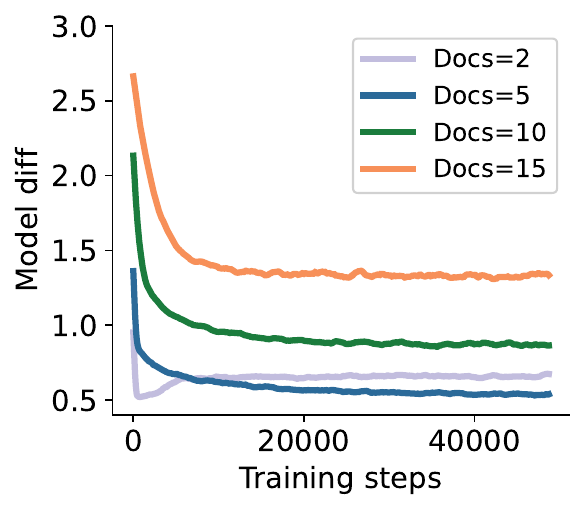}%
    \includegraphics[width=0.24\textwidth]{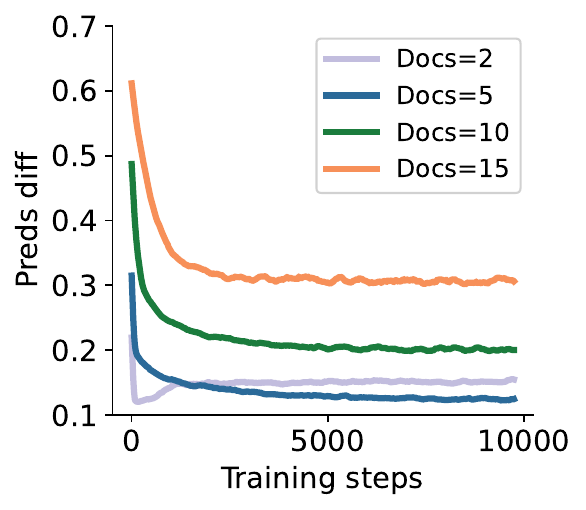}\\[1ex]
    \includegraphics[width=0.24\textwidth]{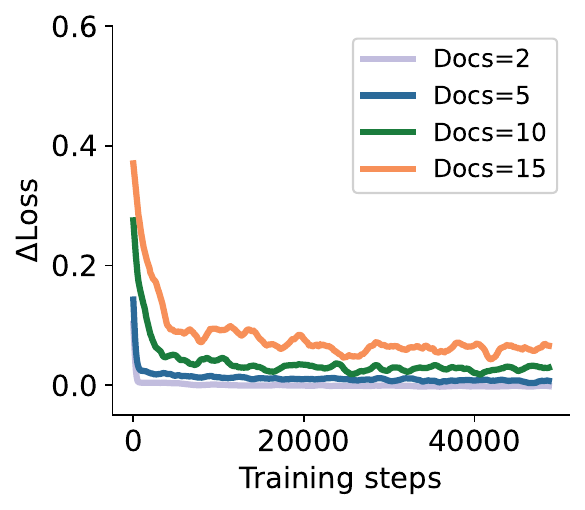}%
    \includegraphics[width=0.24\textwidth]{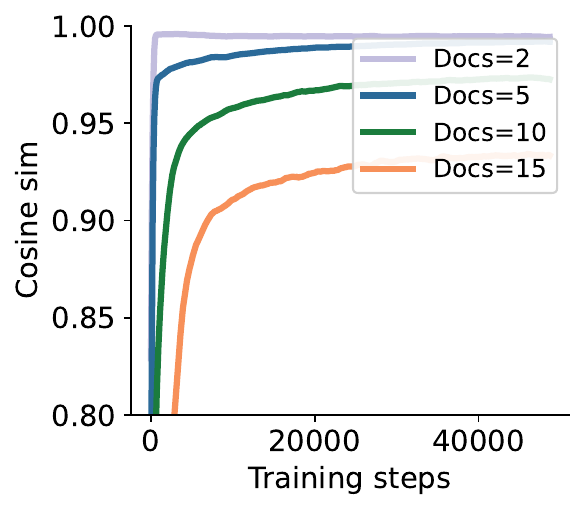}%
    \includegraphics[width=0.24\textwidth]{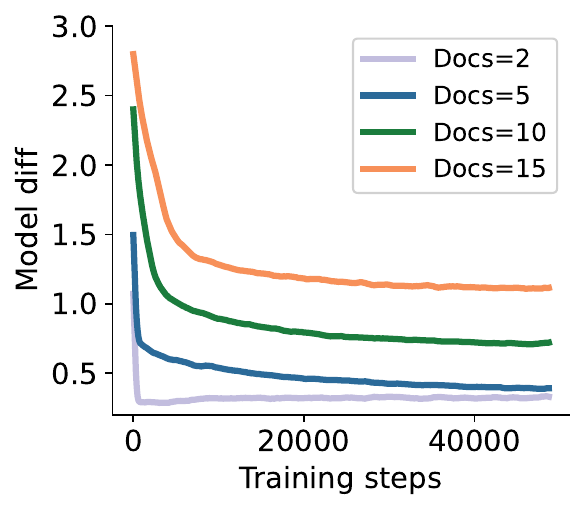}%
    \includegraphics[width=0.24\textwidth]{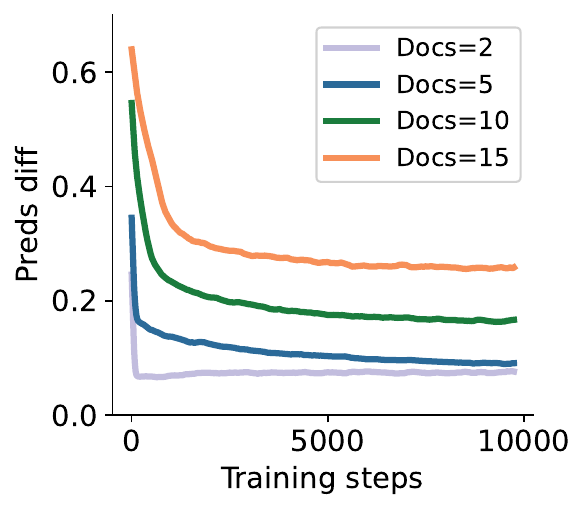}%
    \caption{Stacked-layer agreement under the projection-based retriever. \textit{Top row:} 2-layer model. \textit{Bottom row:} 5-layer model. Columns: (a) loss difference between trained Transformer and constructed gradient-descent predictor, (b) sensitivity cosine, (c) model difference, (d) prediction difference.}
    \label{figure:layers_different_doc_R1}
\end{figure*}

This appendix complements the dot-product analysis of Appendix~\ref{appendix:linear} with the projection-based retriever. Under this interface, the retrieved documents are concatenated with the input tokens, $e_i = (x_i, \mathcal{D}, y_i)$, so each per-document feature is processed jointly with the query through every stacked layer. As the document count grows, the variance of the per-token features grows with it, which amplifies the discrepancy between the trained Transformer and the constructed gradient-descent predictor at any fixed depth. Stacking layers reduces this discrepancy: at depth 5, the loss and prediction differences are uniformly smaller across document counts than at depth 2, and the sensitivity cosine is closer to 1.

Whereas the dot-product sweep in Appendix~\ref{appendix:linear} reports results at 2, 5, 10, and 25 documents, the projection-based sweep is restricted to 2, 5, 10, and 15. The compute cost of stacking concatenated-document tokens grows substantially with retrieval size, and the growing-residual trend is already clear at 15 documents, so the 25-document run is omitted.

\section{Dataset Details}

\paragraph{QA benchmarks (main evaluation).}
The seven question-answering benchmarks used in the main evaluation are all publicly available:
\begin{itemize}
    \item \textbf{Natural Questions (NQ)}~\citep{kwiatkowski2019natural}: open-domain factoid QA over Wikipedia.
    \item \textbf{TriviaQA}~\citep{joshi2017triviaqa}: large-scale trivia question answering with evidence documents.
    \item \textbf{PopQA}~\citep{mallen2023popqa}: popularity-stratified entity-centric QA (held out from training).
    \item \textbf{HotpotQA}~\citep{yang2018hotpotqa}: multi-hop QA with comparison and bridge questions.
    \item \textbf{2WikiMultiHopQA}~\citep{ho20202wiki}: multi-hop questions grounded in Wikipedia article pairs.
    \item \textbf{MuSiQue}~\citep{trivedi2022musique}: compositional multi-hop QA constructed by composing single-hop pairs.
    \item \textbf{Bamboogle}~\citep{press2023bamboogle}: small multi-hop benchmark on long-tail entities (held out from training).
\end{itemize}
For each benchmark we use the standard FlashRAG~\citep{jin2024flashrag} corpus and retrieval splits.
NQ is used to train the source retrieval adapter $W_0^{\mathrm{ret}}$.
NQ, TriviaQA, HotpotQA, 2WikiMultiHopQA, and MuSiQue are used to meta-train the predictor $g_\phi$.
PopQA and Bamboogle are held out from both adapter training and predictor meta-training, and are used only for evaluation.

\paragraph{Synthetic and tabular regression datasets (linear-attention sanity check and normalization analysis).}
\label{con:dataset_detals}
\begin{itemize}
    \item \textbf{California Housing}: Given eight features,
    \texttt{['MedInc', 'HouseAge', 'AveRooms', 'AveBedrms', 'Population', 'AveOccup', 'Latitude', 'Longitude']},
    the task is to predict \texttt{MedHouseVal}. The dataset is split into 16,640 training samples and 2,000 test samples.

    \item \textbf{Bike Sharing}: Using the features
    \texttt{['season', 'yr', 'mnth', 'hr', 'holiday', 'weekday', 'workingday', 'weathersit', 'temp', 'atemp', 'hum', 'windspeed', 'casual', 'registered']},
    the task is to predict \texttt{count}. The dataset contains 15,641 training samples and 1,738 test samples.

    \item  \textbf{Wine Quality}: Given eleven physicochemical features,
\texttt{[fixed acidity, volatile acidity, citric acid, residual sugar, chlorides, free sulfur dioxide, total sulfur dioxide, density, pH, sulphates, alcohol]},
the task is to predict the wine \texttt{quality} (a sensory score ranging from 0 to 10).
The dataset is split into 4,408 training samples and 490 test samples.

\item \textbf{Predict Calorie Expenditure}: Using the features
\texttt{[Gender, Age, Height, Weight, Duration, Heart\_Rate, Body\_Temp]},
the task is to predict the number of \texttt{Calories} expended.
The dataset is split into 13,500 training samples and 1,540 test samples.

\end{itemize}

\section{Implementation Details}
\label{app:impl}

\paragraph{Source retrieval adapter $W_0^{\mathrm{ret}}$.}
We implement the generator-side retrieval interface with LoRA modules on the $\{q,k,v\}$ projections of every Transformer block.
The LoRA rank is $16$, with $\alpha{=}32$ and dropout $0$.
The LLM backbone is kept frozen and loaded in 4-bit NF4 precision.
We train $W_0^{\mathrm{ret}}$ with AdamW using learning rate $10^{-4}$, weight decay $0.01$, gradient clipping $1.0$, and gradient accumulation $4$ for $3{,}000$ steps on RAG-formatted examples from the NQ training split.
Each example is paired with the top-$K_{\mathrm{ret}}{=}5$ retrieved documents from the fixed external retriever.
For each retriever setting, we use the corresponding fixed retrieval cache and keep this cache unchanged across methods.
This adapter serves only as a generator-side evidence-use interface: it does not select documents, but modulates how the frozen generator uses the retrieved evidence.

\paragraph{Predictor $g_\phi$.}
The predictor maps a support context $C=\{(x_i,D_i,y_i)\}_{i=1}^{N}$ with $N{=}3$ demonstrations to a context-conditioned update for the generator-side retrieval interface.
Each demonstration is formatted by concatenating the question, retrieved documents, and gold answer, and is encoded by the frozen LLM equipped with $W_0^{\mathrm{ret}}$.
We take the EOS hidden state $h_i$ of each demonstration and aggregate the support context by mean pooling,
\[
\bar{h}(C)=\frac{1}{N}\sum_{i=1}^{N}h_i .
\]
The pooled representation is passed to a two-layer MLP encoder with hidden dimension $256$ and output dimension $64$, followed by per-layer and per-projection update heads for the $\{q,k,v\}$ LoRA modules.
The update heads output low-rank perturbations with the same LoRA rank as $W_0^{\mathrm{ret}}$, so that $g_\phi(C)$ has the same parameter shape as the base retrieval adapter.
We train $g_\phi$ with AdamW using learning rate $5\times10^{-4}$, weight decay $0.01$, gradient clipping $1.0$, and gradient accumulation $4$ for $3{,}000$ steps.
The predictor is meta-trained on support contexts sampled from NQ, TriviaQA, HotpotQA, 2WikiMultiHopQA, and MuSiQue.
PopQA and Bamboogle are excluded from both adapter training and predictor meta-training, and are used only for held-out evaluation.

\paragraph{Matching objective.}
The predictor is trained to match the autograd-defined inner-GD target for each layer and projection type.
The matching loss uses a cosine term for update direction and a log-magnitude term for update scale, as defined in Section~\ref{sec:llm_predictor}.
We set the magnitude weight to $\lambda{=}0.1$ throughout.
No downstream answer loss is applied when training $g_\phi$ for the main \textsc{RAG-GD} results.

\paragraph{Inner GD target.}
For each support context, the supervision target is computed by running $K$ steps of SGD on the $N{=}3$ demonstrations, starting from $W_0^{\mathrm{ret}}$.
The inner-loop learning rate is $\eta{=}10^{-2}$, and we evaluate $K\in\{1,5,10\}$.
Gradients are taken only with respect to the LoRA parameters of the generator-side retrieval interface; the external retriever and the LLM backbone remain fixed.

\paragraph{Compute.}
All experiments are run on NVIDIA A100 80GB GPUs.
A single training run for $W_0^{\mathrm{ret}}$ takes approximately $50$--$60$ minutes.
Training $g_\phi$ takes approximately $1.5$ hours for $K{=}1$, $4$ hours for $K{=}5$, and $7$ hours for $K{=}10$.
The full result table requires approximately $50$ A100-hours.

\section{Algorithm and Inference Procedure for \textsc{RAG-GD}}
\label{appendix:algorithm}

Algorithm~\ref{alg:forward_only_2e} summarizes the deployment-time procedure of \textsc{RAG-GD}: build a RAG-formatted support context, predict the retrieval-interface update with a single forward pass through $g_\phi$, and generate with the perturbed interface. No backward pass through the LLM is required at deployment.

\begin{algorithm}[h]
\DontPrintSemicolon
\caption{\textsc{RAG-GD}: Forward-Only Retrieval-Interface Adaptation}
\label{alg:forward_only_2e}
\KwIn{External retriever $R$, frozen LLM $f$, source adapter $W_0^{\mathrm{ret}}$, predictor $g_\phi$, support size $N$, query $q$}
\KwOut{Generated answer $\hat y_q$}

\textbf{// Phase 1: Build RAG-formatted support context}\;
$C \leftarrow \emptyset$\;
\For{$i=1,\ldots,N$}{
    Sample support pair $(x_i,y_i)$ from the task support pool\;
    $\mathcal{D}_i \leftarrow R(x_i)$ \tcp*[r]{retrieve top-$k$ documents}
    $C \leftarrow C \cup \{(x_i,\mathcal{D}_i,y_i)\}$\;
}

\textbf{// Phase 2: Predict retrieval-interface update}\;
\For{$(x_i,\mathcal{D}_i,y_i)\in C$}{
    $h_i \leftarrow f_{W_0^{\mathrm{ret}}}(x_i,\mathcal{D}_i,y_i)_{\mathrm{EOS}}$\;
}
$\bar h(C) \leftarrow \frac{1}{N}\sum_{i=1}^{N}h_i$\;
$\widetilde{\Delta W}(C) \leftarrow g_\phi(\bar h(C))$\;

\textbf{// Phase 3: Generate with adapted interface}\;
$\mathcal{D}_q \leftarrow R(q)$\;
$\hat y_q \leftarrow f_{W_0^{\mathrm{ret}}+\widetilde{\Delta W}(C)}(q,\mathcal{D}_q)$\;
\Return $\hat y_q$\;
\end{algorithm}

\section{Full Per-Method Comparison on QA Benchmarks}
\label{appendix:full_qa_table}

\begin{figure*}[!htbp]
\centering
\includegraphics[width=\textwidth]{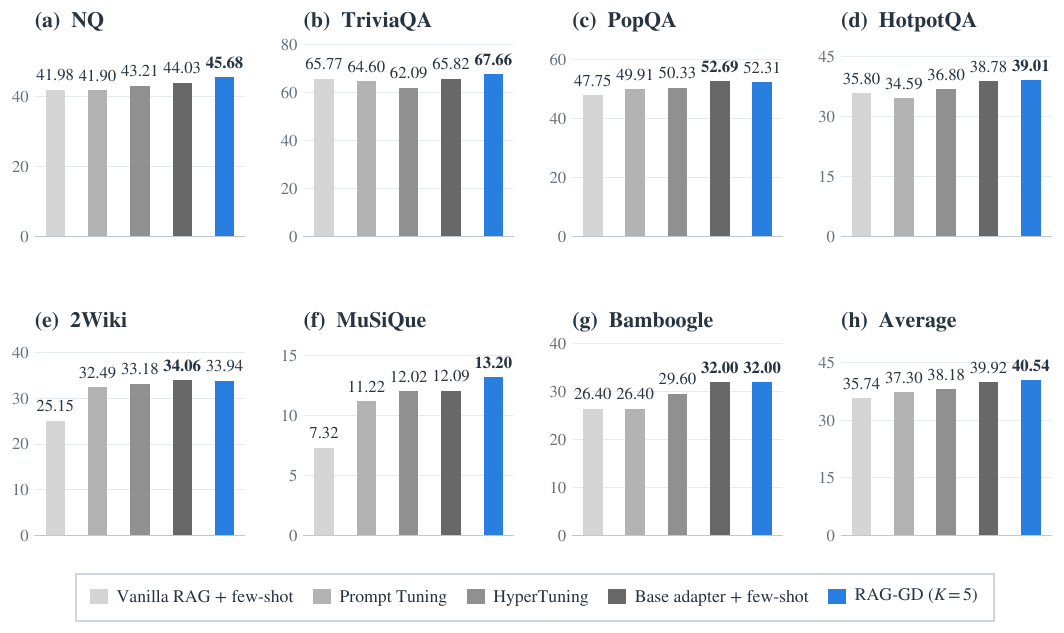}
\caption{Context-conditioned baseline comparison on Llama 3.1-8B-Instruct with E5 retrieval. Panels (a)--(g) show exact match (EM) on the seven benchmarks; (h) shows their unweighted average. Values above bars report EM, with the best score in each panel shown in bold, including ties. Each panel uses its own vertical scale starting at zero.}
\label{fig:llama_e5_baselines}
\end{figure*}

We report the per-benchmark results in two tables and one figure. Table~\ref{tab:full_qa_shared} reports the methods that we ran on \emph{both} Qwen 2.5-7B-Instruct and Llama 3.1-8B-Instruct: Query Only, Vanilla RAG, Base adapter, and \textsc{RAG-GD} at $K\in\{1,5,10\}$. Table~\ref{tab:full_qa_qwen} reports the additional context-conditioned baselines on Qwen: Vanilla RAG\,+\,few-shot, Base adapter\,+\,few-shot, Prompt Tuning, HyperTuning, and TT-SGD at $K{=}5$. Figure~\ref{fig:llama_e5_baselines} reports four of these baselines on Llama with E5 retrieval, alongside \textsc{RAG-GD} ($K{=}5$). Together they complement the slim main-text Table~\ref{tab:main} and the per-family aggregates in Figures~\ref{fig:family_ktrend} and~\ref{fig:efficiency}. \textbf{HyperTuning}~\citep{phang2023hypertuning} uses the same predictor architecture as \textsc{RAG-GD} but supervises against a downstream task loss instead of the SGD-update target, so the contrast between the HyperTuning rows in Table~\ref{tab:full_qa_qwen} and the \textsc{RAG-GD} ($K{=}5$) rows for Qwen in Table~\ref{tab:full_qa_shared} isolates the choice of supervision signal. \textbf{TT-SGD} performs $K{=}5$ inner gradient-descent steps at test time and serves as a reference for what test-time gradient adaptation would achieve at the same backbone and retriever. \textbf{+ few-shot} variants concatenate the support demonstrations into the prompt under the corresponding base configuration.

\begin{table*}[!htb]
\centering
\caption{Methods run on both backbones. Per-benchmark exact match (EM) and F1 for Query Only, Vanilla RAG, Base adapter, and \textsc{RAG-GD} at $K\in\{1,5,10\}$.}
\scriptsize
\setlength{\tabcolsep}{2.8pt}
\renewcommand{\arraystretch}{1}
\resizebox{\textwidth}{!}{
\begin{tabular}{llcccccccccccccccc}
\toprule
\multirow{3}{*}{\textbf{Method}} & \multirow{3}{*}{\textbf{Retriever}}
& \multicolumn{6}{c}{\textbf{Single-Hop QA}}
& \multicolumn{8}{c}{\textbf{Multi-Hop QA}}
& \multicolumn{2}{c}{\textbf{Avg.}} \\
\cmidrule(lr){3-8} \cmidrule(lr){9-16} \cmidrule(lr){17-18}
& & \multicolumn{2}{c}{\textbf{NQ}} & \multicolumn{2}{c}{\textbf{TriviaQA}} & \multicolumn{2}{c}{\textbf{PopQA}}
& \multicolumn{2}{c}{\textbf{HotpotQA}} & \multicolumn{2}{c}{\textbf{2Wiki}} & \multicolumn{2}{c}{\textbf{MuSiQue}} & \multicolumn{2}{c}{\textbf{Bamboogle}}
& & \\
\cmidrule(lr){3-4} \cmidrule(lr){5-6} \cmidrule(lr){7-8} \cmidrule(lr){9-10} \cmidrule(lr){11-12} \cmidrule(lr){13-14} \cmidrule(lr){15-16}
& & EM & F1 & EM & F1 & EM & F1 & EM & F1 & EM & F1 & EM & F1 & EM & F1 & EM & F1 \\
\midrule
\multicolumn{18}{c}{\textbf{Qwen 2.5-7B}} \\
\midrule
Query Only & --
& 15.95 & 24.28 & 43.33 & 49.51 & 16.02 & 19.76 & 18.40 & 25.39 & 23.91 & 28.12 & 3.80 & 10.57 & 11.20 & 18.02 & 18.94 & 25.09 \\
\midrule
\multirow{2}{*}{Vanilla RAG}
& BM25 & 27.61 & 36.66 & 58.24 & 65.77 & 28.84 & 33.18 & 31.28 & 41.25 & 27.87 & 33.24 & 5.87 & 13.05 & 10.40 & 21.16 & 27.16 & 34.90 \\
& E5   & 39.16 & 50.03 & 62.99 & 70.80 & 44.03 & 50.21 & 32.45 & 42.21 & 25.48 & 31.43 & 5.79 & 12.77 & 18.40 & 26.74 & 32.61 & 40.60 \\
\midrule
\multirow{2}{*}{Base adapter}
& BM25 & 32.57 & 41.45 & 60.11 & 67.93 & 31.55 & 35.63 & 32.31 & 43.45 & 28.22 & 34.13 & 6.41 & 15.46 & 16.80 & 26.01 & 29.71 & 37.72 \\
& E5   & 41.77 & 51.22 & 63.31 & 71.62 & 47.05 & 51.82 & 33.78 & 44.64 & 27.89 & 33.97 & 6.95 & 15.76 & 18.40 & 28.78 & 34.16 & 42.54 \\
\midrule
\rowcolor{ragtableblue!12}
& BM25 & 33.88 & 43.20 & 63.02 & 70.66 & 33.25 & 37.73 & 35.25 & 46.79 & 28.84 & 34.59 & 9.14 & 19.62 & 22.40 & 33.06 & 32.25 & 40.81 \\
\rowcolor{ragtableblue!12}
\multirow{-2}{*}{\textcolor{ragtabletext}{\textbf{\textsc{RAG-GD} ($K{=}1$)}}}
& E5   & 42.49 & 52.32 & 65.61 & 73.51 & 48.41 & 52.93 & 35.34 & 46.94 & 29.66 & 35.50 & 8.93 & 19.17 & 24.00 & 34.18 & 36.35 & 44.94 \\
\midrule
\rowcolor{ragtableblue!12}
& BM25 & 34.46 & 43.54 & 63.27 & 70.69 & 33.22 & 37.69 & 35.54 & 47.14 & 28.86 & 34.48 & 9.26 & 19.73 & 22.40 & 32.85 & 32.43 & 40.87 \\
\rowcolor{ragtableblue!12}
\multirow{-2}{*}{\textcolor{ragtabletext}{\textbf{\textsc{RAG-GD} ($K{=}5$)}}}
& E5   & 42.91 & 52.71 & 65.98 & 73.60 & 48.12 & 52.61 & 35.54 & 47.00 & 29.67 & 35.47 & 9.14 & 19.26 & 25.60 & 35.12 & 36.71 & 45.11 \\
\midrule
\rowcolor{ragtableblue!12}
& BM25 & 34.57 & 43.72 & 63.26 & 70.57 & 33.11 & 37.57 & 35.45 & 47.07 & 28.64 & 34.32 & 9.35 & 19.58 & 22.40 & 31.92 & 32.40 & 40.68 \\
\rowcolor{ragtableblue!12}
\multirow{-2}{*}{\textcolor{ragtabletext}{\textbf{\textsc{RAG-GD} ($K{=}10$)}}}
& E5   & 42.46 & 52.14 & 65.93 & 73.49 & 47.96 & 52.43 & 35.69 & 47.05 & 29.43 & 35.14 & 8.90 & 19.09 & 26.40 & 34.67 & 36.68 & 44.86 \\
\midrule
\multicolumn{18}{c}{\textbf{Llama 3.1-8B}} \\
\midrule
Query Only & --
& 22.46 & 32.51 & 52.67 & 59.79 & 20.63 & 25.09 & 18.31 & 25.71 & 26.39 & 31.04 & 3.81 & 9.51 & 6.40 & 12.88 & 21.52 & 28.08 \\
\midrule
\multirow{2}{*}{Vanilla RAG}
& BM25 & 31.41 & 40.95 & 60.43 & 68.35 & 31.08 & 35.43 & 31.92 & 42.46 & 26.07 & 31.82 & 5.75 & 12.44 & 14.40 & 22.93 & 28.72 & 36.34 \\
& E5   & 40.72 & 52.23 & 64.42 & 72.58 & 45.85 & 51.48 & 32.78 & 43.19 & 23.44 & 29.46 & 6.04 & 12.25 & 24.80 & 32.12 & 34.01 & 41.90 \\
\midrule
\multirow{2}{*}{Base adapter}
& BM25 & 38.47 & 49.42 & 62.66 & 72.46 & 37.80 & 42.29 & 37.35 & 50.41 & 33.66 & 39.55 & 11.46 & 21.98 & 29.60 & 41.29 & 35.86 & 45.34 \\
& E5   & 43.46 & 54.60 & 63.90 & 74.05 & 51.69 & 55.74 & 37.31 & 49.90 & 33.45 & 39.45 & 11.83 & 22.07 & 30.40 & 40.48 & 38.86 & 48.04 \\
\midrule
\rowcolor{ragtableblue!12}
& BM25 & 39.58 & 49.60 & 66.13 & 74.13 & 37.89 & 42.38 & 38.53 & 50.67 & 33.72 & 39.61 & 12.25 & 22.48 & 30.40 & 41.69 & 36.93 & 45.79 \\
\rowcolor{ragtableblue!12}
\multirow{-2}{*}{\textcolor{ragtabletext}{\textbf{\textsc{RAG-GD} ($K{=}1$)}}}
& E5   & 45.10 & 55.19 & 67.63 & 75.98 & 52.38 & 56.61 & 38.27 & 50.09 & 33.44 & 39.47 & 12.32 & 22.52 & 30.40 & 40.72 & 39.93 & 48.65 \\
\midrule
\rowcolor{ragtableblue!12}
& BM25 & 40.22 & 50.01 & 66.13 & 74.28 & 37.81 & 42.19 & 38.99 & 51.14 & 34.15 & 40.04 & 12.54 & 22.61 & 28.00 & 39.61 & 36.83 & 45.70 \\
\rowcolor{ragtableblue!12}
\multirow{-2}{*}{\textcolor{ragtabletext}{\textbf{\textsc{RAG-GD} ($K{=}5$)}}}
& E5   & 45.68 & 55.84 & 67.66 & 76.07 & 52.31 & 56.48 & 39.01 & 50.94 & 33.94 & 39.83 & 13.20 & 23.47 & 32.00 & 42.08 & 40.54 & 49.24 \\
\midrule
\rowcolor{ragtableblue!12}
& BM25 & 40.28 & 50.29 & 65.93 & 74.07 & 37.73 & 42.10 & 39.10 & 51.20 & 34.26 & 40.08 & 12.62 & 22.86 & 28.80 & 39.92 & 36.96 & 45.79 \\
\rowcolor{ragtableblue!12}
\multirow{-2}{*}{\textcolor{ragtabletext}{\textbf{\textsc{RAG-GD} ($K{=}10$)}}}
& E5   & 45.48 & 55.51 & 67.71 & 76.03 & 52.38 & 56.49 & 39.10 & 50.96 & 34.03 & 39.79 & 13.41 & 23.41 & 32.00 & 42.83 & 40.59 & 49.29 \\
\bottomrule
\end{tabular}
}
\vspace{4pt}
\label{tab:full_qa_shared}
\end{table*}

\begin{table*}[!htb]
\centering
\caption{Additional context-conditioned baselines on Qwen 2.5-7B. For comparison anchors (Qwen Base adapter and \textsc{RAG-GD} at the same retriever and benchmark), see Table~\ref{tab:full_qa_shared}.}
\scriptsize
\setlength{\tabcolsep}{2.8pt}
\renewcommand{\arraystretch}{1}
\resizebox{\textwidth}{!}{
\begin{tabular}{llcccccccccccccccc}
\toprule
\multirow{3}{*}{\textbf{Method}} & \multirow{3}{*}{\textbf{Retriever}}
& \multicolumn{6}{c}{\textbf{Single-Hop QA}}
& \multicolumn{8}{c}{\textbf{Multi-Hop QA}}
& \multicolumn{2}{c}{\textbf{Avg.}} \\
\cmidrule(lr){3-8} \cmidrule(lr){9-16} \cmidrule(lr){17-18}
& & \multicolumn{2}{c}{\textbf{NQ}} & \multicolumn{2}{c}{\textbf{TriviaQA}} & \multicolumn{2}{c}{\textbf{PopQA}}
& \multicolumn{2}{c}{\textbf{HotpotQA}} & \multicolumn{2}{c}{\textbf{2Wiki}} & \multicolumn{2}{c}{\textbf{MuSiQue}} & \multicolumn{2}{c}{\textbf{Bamboogle}}
& & \\
\cmidrule(lr){3-4} \cmidrule(lr){5-6} \cmidrule(lr){7-8} \cmidrule(lr){9-10} \cmidrule(lr){11-12} \cmidrule(lr){13-14} \cmidrule(lr){15-16}
& & EM & F1 & EM & F1 & EM & F1 & EM & F1 & EM & F1 & EM & F1 & EM & F1 & EM & F1 \\
\midrule
\multirow{2}{*}{Vanilla RAG + few-shot}
& BM25 & 27.72 & 36.80 & 58.41 & 65.98 & 28.80 & 33.18 & 31.73 & 41.43 & 26.27 & 32.05 & 5.88 & 13.58 & 11.20 & 20.27 & 27.14 & 34.76 \\
& E5   & 38.78 & 49.75 & 63.14 & 70.94 & 43.82 & 50.00 & 32.28 & 42.05 & 24.20 & 30.80 & 5.99 & 13.21 & 16.80 & 25.42 & 32.14 & 40.31 \\
\midrule
\multirow{2}{*}{Base adapter + few-shot}
& BM25 & 34.07 & 43.18 & 62.26 & 69.97 & 32.94 & 37.65 & 34.65 & 46.18 & 28.91 & 34.56 & 9.06 & 19.66 & 19.20 & 30.19 & 31.58 & 40.20 \\
& E5   & 41.52 & 51.25 & 65.03 & 73.11 & 47.85 & 52.38 & 34.51 & 46.08 & 29.81 & 35.70 & 8.44 & 18.67 & 26.40 & 36.08 & 36.22 & 44.75 \\
\midrule
\multirow{2}{*}{Prompt Tuning}
& BM25 & 28.94 & 40.81 & 58.47 & 69.11 & 32.42 & 37.14 & 32.43 & 44.98 & 32.18 & 38.54 & 7.36 & 17.79 & 22.40 & 33.15 & 30.60 & 40.22 \\
& E5   & 35.92 & 48.16 & 61.09 & 71.73 & 47.10 & 51.96 & 32.72 & 45.14 & 31.60 & 38.04 & 7.28 & 17.22 & 26.40 & 32.98 & 34.59 & 43.60 \\
\midrule
\multirow{2}{*}{HyperTuning}
& BM25 & 33.62 & 43.21 & 61.38 & 69.90 & 33.71 & 37.97 & 34.55 & 46.34 & 31.87 & 37.81 & 7.52 & 17.60 & 19.20 & 31.69 & 31.69 & 40.65 \\
& E5   & 40.72 & 50.51 & 64.19 & 72.57 & 48.88 & 53.25 & 34.76 & 46.16 & 31.47 & 37.44 & 6.95 & 16.74 & 24.00 & 32.83 & 35.85 & 44.21 \\
\midrule
\multirow{2}{*}{TT-SGD ($K{=}5$)}
& BM25 & 34.37 & 43.32 & 63.32 & 70.75 & 32.94 & 37.63 & 35.16 & 47.13 & 28.76 & 34.53 & 8.77 & 19.51 & 23.20 & 33.26 & 32.36 & 40.88 \\
& E5   & 42.52 & 52.14 & 65.50 & 73.45 & 47.95 & 52.51 & 35.62 & 47.43 & 29.67 & 35.59 & 8.93 & 19.13 & 25.60 & 34.88 & 36.54 & 45.02 \\
\bottomrule
\end{tabular}
}
\vspace{4pt}
\label{tab:full_qa_qwen}
\end{table*}

We evaluate four additional baselines on Llama-3.1-8B-Instruct with E5 retrieval, following the same protocol as the Qwen experiments (Figure~\ref{fig:llama_e5_baselines}).
\textsc{RAG-GD} outperforms HyperTuning on all seven benchmarks, achieving an average EM of 40.54 compared with 38.18 for HyperTuning.
It also exceeds Base adapter + few-shot, the strongest of these four baselines, by 0.62 EM points on average, consistent with the 0.49-point improvement on Qwen.

%%%%%%%%%%%%%%%%%%%%%%%%%%%%%%%%%%%%%%%%%%%%%%%%%%%%%%%%%%%%

\section{Additional Evaluation on Qwen3-14B}
\label{appendix:qwen3}

We additionally evaluate Qwen3-14B with E5 retrieval under the same protocol as the Qwen 2.5-7B experiments on five benchmarks: NQ, TriviaQA, PopQA, HotpotQA, and Bamboogle.
Table~\ref{tab:qwen3_e5} reports EM and F1, with averages computed over these five benchmarks.

\begin{table*}[!htb]
\centering
\caption{Additional RAG-GD evaluation on Qwen3-14B with E5 retrieval. Avg. is the unweighted mean over the five evaluated benchmarks. Bold values mark the best result per column.}
\label{tab:qwen3_e5}
\scriptsize
\setlength{\tabcolsep}{2.8pt}
\renewcommand{\arraystretch}{1}
\resizebox{0.85\textwidth}{!}{
\begin{tabular}{lcccccccccccc}
\toprule
\multirow{3}{*}{\textbf{Method}}
& \multicolumn{6}{c}{\textbf{Single-Hop QA}}
& \multicolumn{4}{c}{\textbf{Multi-Hop QA}}
& \multicolumn{2}{c}{\textbf{Avg.}} \\
\cmidrule(lr){2-7} \cmidrule(lr){8-11} \cmidrule(lr){12-13}
& \multicolumn{2}{c}{\textbf{NQ}} & \multicolumn{2}{c}{\textbf{TriviaQA}} & \multicolumn{2}{c}{\textbf{PopQA}}
& \multicolumn{2}{c}{\textbf{HotpotQA}} & \multicolumn{2}{c}{\textbf{Bamboogle}} & & \\
\cmidrule(lr){2-3} \cmidrule(lr){4-5} \cmidrule(lr){6-7} \cmidrule(lr){8-9} \cmidrule(lr){10-11}
& EM & F1 & EM & F1 & EM & F1 & EM & F1 & EM & F1 & EM & F1 \\
\midrule
Vanilla RAG & 39.03 & 51.81 & 63.59 & 71.48 & 39.79 & 51.98 & 37.81 & 49.14 & 23.20 & 31.05 & 40.68 & 51.09 \\
\midrule
Vanilla RAG + few-shot & 41.31 & 52.60 & 65.41 & 72.18 & 42.51 & 53.41 & 38.10 & 49.48 & 22.40 & 32.21 & 41.95 & 51.98 \\
\midrule
Base adapter & 44.30 & 54.84 & 64.92 & 73.53 & 49.38 & 57.16 & 41.09 & 53.50 & 28.80 & 44.17 & 45.70 & 56.64 \\
\midrule
Base adapter + few-shot & 44.62 & 55.35 & 65.11 & 73.51 & 50.39 & 57.77 & 40.58 & 52.80 & 29.60 & 44.61 & 46.06 & 56.81 \\
\midrule
\rowcolor{ragtableblue!12}
\textcolor{ragtabletext}{\textbf{\textsc{RAG-GD} ($K{=}5$)}} & \textbf{45.24} & \textbf{55.62} & \textbf{66.42} & \textbf{74.94} & \textbf{51.80} & \textbf{57.83} & \textbf{42.11} & \textbf{54.66} & \textbf{32.00} & \textbf{46.65} & \textbf{47.51} & \textbf{57.94} \\
\bottomrule
\end{tabular}
}
\end{table*}

\textsc{RAG-GD} ($K{=}5$) outperforms Base adapter + few-shot on all five benchmarks in both EM and F1, improving average performance by 1.45 EM and 1.13 F1 points.
On the same five-benchmark subset, the corresponding EM margin on Qwen 2.5-7B is 0.57 points.

On Qwen3-14B, few-shot prompting raises Vanilla RAG's average EM from 40.68 to 41.95, but raises Base adapter's average EM only from 45.70 to 46.06.
On Qwen 2.5-7B, the corresponding changes on the same subset are $-0.44$ and $+2.20$ EM points, respectively.
These results show that the benefit of \textsc{RAG-GD} persists on Qwen3-14B beyond few-shot prompting.

\section{Reference-Based Semantic Evaluation}
\label{appendix:semantic_eval}

\begin{table}[!htbp]
\centering
\caption{EM, F1, and reference-based LLM-judge accuracy for Qwen 2.5-7B-Instruct with E5 retrieval. All scores are percentages macro-averaged over the same seven benchmarks. Bold values mark the best result per column.}
\label{tab:semantic_eval}
\footnotesize
\setlength{\tabcolsep}{2.5pt}
\renewcommand{\arraystretch}{1}
\begin{tabularx}{0.7\linewidth}{l>{\centering\arraybackslash}X>{\centering\arraybackslash}Xc}
\toprule
\textbf{Method} & \textbf{EM} & \textbf{F1} & \makecell{\textbf{LLM-judge}\\\textbf{accuracy}} \\
\midrule
Vanilla RAG & 32.61 & 40.60 & 41.25 \\
\midrule
Base adapter & 34.16 & 42.54 & 43.88 \\
\midrule
Base adapter + few-shot & 36.22 & 44.75 & 45.38 \\
\midrule
HyperTuning & 35.85 & 44.21 & 44.91 \\
\midrule
\rowcolor{ragtableblue!12}
\textcolor{ragtabletext}{\textbf{\textsc{RAG-GD} ($K{=}5$)}} & \textbf{36.71} & \textbf{45.11} & \textbf{46.04} \\
\bottomrule
\end{tabularx}
\end{table}

To complement EM and F1, we evaluate the Qwen 2.5-7B-Instruct predictions with E5 retrieval using Qwen2.5-32B-Instruct as a reference-based judge with deterministic decoding.
Given the question, gold answer, and prediction, the judge assesses whether the prediction is semantically equivalent to the reference.
LLM-judge accuracy is the fraction of predictions judged equivalent, macro-averaged across the same seven benchmarks.

As shown in Table~\ref{tab:semantic_eval}, LLM-judge scores are 8.64--9.72 percentage points higher than EM for all evaluated methods.
The method ranking is unchanged across the three metrics, with \textsc{RAG-GD} achieving the highest score under each metric.

Examples of predictions accepted by the judge but rejected by EM include ``International Whaling Commission'' for ``The International Whaling Commission or IWC'', ``documentaries'' for ``documentary'', and ``health points'' for ``hit points or health points''.
These examples illustrate mismatches involving aliases, inflections, and reference formatting.

\section{Additional Related Work}
\label{sec:additional-related-work}

\paragraph{Retrieval and Knowledge Integration.}
Retrieval-augmented generation (RAG) incorporates external
knowledge into language models to support knowledge-intensive
tasks~\citep{lewis2020retrieval}. Its applications have expanded
to specialized domains and diverse prediction tasks.
RT~\citep{li2024rt} combines retrieved demonstrations with
chain-of-thought reasoning for few-shot medical named entity
recognition. BiomedRAG~\citep{li2025biomedrag} integrates
biomedical evidence into language models for healthcare
applications, while CancerLLM~\citep{li2026cancerllm}
uses retrieval-based example selection for cancer phenotyping
and diagnosis generation. Prior studies have also examined
the robustness and self-awareness of retrieval-augmented
models under noisy and heterogeneous evidence
~\citep{li2025benchmarking}. Beyond external knowledge
sources, K2K~\citep{li2026efficient} enables internal
key-based knowledge access through activation-guided
probing and cross-attentive reranking, providing an
alternative mechanism for incorporating task-relevant
knowledge.

\paragraph{Structured Knowledge Retrieval and Reasoning.}
Beyond unstructured text, structured knowledge provides
relational evidence that can support knowledge-intensive
retrieval and downstream reasoning.
Kiest~\citep{li2023understand} retrieves relevant entities
and attributes from knowledge graphs to facilitate entity
state tracking, while CTLP~\citep{li2024condensed}
captures multi-hop relational paths through condensed
transition graphs for zero-shot link prediction.
RiTeK~\citep{huang2026ritek} introduces a benchmark for
complex reasoning over textual knowledge graphs and
evaluates retrieval methods for identifying relational
evidence. GLEN-Bench~\citep{huang2026glen} further
investigates graph-language reasoning over structured
knowledge. Related efforts explore the use of graph
structures to support downstream reasoning processes.
EvolveRouter~\citep{huang2026evolverouter} leverages
graph-based relationships for adaptive agent routing
and prompt optimization, while
CAGE-CAL~\citep{huang2026counterfactual} models
inter-agent dependencies through counterfactual graphs
to improve reasoning reliability.
Together, these studies highlight the complementary
roles of structured evidence retrieval and relational
reasoning in knowledge-intensive tasks.

\paragraph{Interactive Retrieval and Learning.}
Recent research has moved beyond fixed retrieval pipelines
toward systems that dynamically adjust evidence acquisition
and reasoning behavior. DeepRAG~\citep{guan2026deeprag}
introduces sequential retrieval decisions that determine
when additional information should be acquired.
RIVO~\citep{li2026rivo} improves long-horizon research
through lightweight, state-conditioned interventions
that redirect agent trajectories without retraining
the underlying model. RICE-PO~\citep{li2026rice}
investigates policy optimization for interactive retrieval
agents by deriving localized credit signals from executable
retrieval actions and counterfactual interactions.
These methods address different aspects of adaptive
retrieval, including search decisions, trajectory
control, and learning from retrieval feedback.

While prior studies investigate how knowledge is retrieved,
represented, or incorporated into language models, a
related question is whether retrieved context can also
serve as a signal for adapting the generator itself.
In this work, we study this question from an
in-context optimization perspective. Rather than
modifying the retrieval process or updating the
entire language model, we examine how contextual
examples can induce changes in the interaction
between queries and retrieved evidence.
We establish a gradient-descent correspondence
in a linearized RAG setting and use this perspective
to develop a forward-only, context-conditioned
adaptation mechanism for frozen RAG LLMs.

\section{Use of Large Language Models}

Large language models (LLMs) were used to assist with language polishing and minor grammatical editing of this manuscript.
Qwen2.5-32B-Instruct was also used for the reference-based semantic evaluation described in Appendix~\ref{appendix:semantic_eval}.

\end{document}